%% file: main.tex
\documentclass[lettersize,journal]{IEEEtran}
\usepackage{amsmath,amsfonts}
\usepackage{algorithm}
\usepackage{array}
\usepackage[caption=false,font=normalsize,labelfont=sf,textfont=sf]{subfig}
\usepackage{textcomp}
\usepackage{stfloats}
\usepackage{url}
\usepackage{verbatim}
\usepackage{graphicx}
\usepackage{cite}
\usepackage{hyperref}

\usepackage{mathtools}
\usepackage{breqn}
\usepackage{graphicx}
\usepackage{amsfonts}
\usepackage{amsmath}
\usepackage{rotating}
\usepackage{dsfont}
\usepackage{multirow}
\usepackage{arydshln}
\usepackage[dvipsnames]{xcolor}
\usepackage{colortbl}
\usepackage{array, makecell}
\usepackage{stmaryrd}
\definecolor{lightgray}{rgb}{0.9, 0.9, 0.9}
\definecolor{color1}{RGB}{196, 164, 132} % brown
\definecolor{color2}{RGB}{30,144,255} % blue
\definecolor{color3}{RGB}{255, 16, 240} % pink
\definecolor{color4}{RGB}{0, 163, 108} % yellow

\usepackage{nicematrix}
\usepackage{algpseudocode}

\usepackage{tabularx, ragged2e}
\definecolor{dodgerblue}{RGB}{30,144,255}

\usepackage{pgfplots}   
\usepgfplotslibrary{colorbrewer}
\newcolumntype{C}{>{\Centering\arraybackslash}X} % centered "X" column

\usepackage{pgfplots}
\usepackage{pgfplotstable}
\pgfplotsset{compat=newest}

\definecolor{mycolor}{rgb}{0.12156862745098039, 0.4666666666666667, 0.7058823529411765} % tab:blue matplotlib
\definecolor{lightgray}{rgb}{0.8, 0.8, 0.8}
\definecolor{lightblue}{RGB}{30,144,255}
\definecolor{aliceblue}{rgb}{0.97, 0.99, 1.0}
\definecolor{dodgerblue}{RGB}{30,144,255}

\usepackage{amsthm}
\newtheorem{lemma}{Lemma}

\newtheorem{proposition}{Proposition}

\usepackage{stackengine} 

\usepackage{tikz} 
\usetikzlibrary{shapes, shadows, arrows}
\usetikzlibrary{positioning}
\tikzset{mynode/.style={align=center,text width=3cm}} 
\SetSymbolFont{stmry}{bold}{U}{stmry}{m}{n}

\begin{document}

\title{Linear Combinations of Patches are Unreasonably Effective for Single-Image Denoising}

\author{S\'ebastien Herbreteau and Charles Kervrann,~\IEEEmembership{Member,~IEEE,}
        % <-this % stops a space
%\thanks{This paper was produced by the IEEE Publication Technology Group. They are in Piscataway, NJ.}% <-this % stops a space

\thanks{Sébastien Herbreteau and Charles Kervrann are with Centre Inria de l'Université de Rennes and Inserm U1143, Institut Curie, PSL University, France (e-mail: sebastien.herbreteau@inria.fr and charles.kervrann@inria.fr).}}

%\thanks{Manuscript received April 19, 2021; revised August 16, 2021.}}

\markboth{Accepted for publication in IEEE Transactions on Image Processing. DOI 10.1109/TIP.2024.3436651.}%
{Shell \MakeLowercase{\textit{et al.}}: A Sample Article Using IEEEtran.cls for IEEE Journals}


\setlength{\fboxsep}{0pt}%
\setlength{\fboxrule}{0pt}%
  
\IEEEpubid{\fbox{%
    \parbox{1.1\textwidth}{%
        \tiny This article has been accepted for publication in IEEE Transactions on Image Processing. Citation information: DOI 10.1109/TIP.2024.3436651. \copyright 2024 IEEE. Personal use of this material is permitted. Permission from IEEE must be obtained for all other uses, in any current or future media, including reprinting/republishing this material for advertising or promotional purposes, creating new collective works, for resale or redistribution to servers or lists, or reuse of any copyrighted component of this work in other works.
    }%
}}


\maketitle

%ADD: This article has been accepted for publication in IEEE Transactions on Image Processing. This is the author's version which has not been fully edited and content may change prior to final publication. Citation information: DOI 10.1109/TIP.2024.3436651

\input{abstract}

\input{introduction}

\input{section2}

\input{section3}

\input{section4}

\input{section5}

\input{conclusion}

\appendix
\input{appendix}

\input{acknowlegdment}

\bibliography{bib.bib}

\end{document}

%% file: abstract.tex
\begin{abstract}
In the past decade, deep neural networks have revolutionized image denoising in achieving significant accuracy improvements by learning on datasets composed of  noisy/clean image pairs. However, this strategy is extremely dependent on training data quality, which is a well-established weakness. To alleviate the requirement to learn image priors externally, single-image (\textit{a.k.a.}, self-supervised or zero-shot) methods perform denoising solely based on the analysis of the input noisy image without external dictionary or training dataset. This work investigates the effectiveness of linear combinations of patches for denoising under this constraint. Although conceptually very simple, we show that linear combinations of patches are enough to achieve state-of-the-art performance. The proposed parametric approach relies on quadratic risk approximation via multiple pilot images to guide the estimation of the combination weights. Experiments on images corrupted  artificially with Gaussian noise  as well as on real-world  noisy images demonstrate that our method is on par with the very best single-image denoisers, outperforming the recent  neural network-based techniques, while being much faster and fully interpretable.
\end{abstract}


%% file: introduction.tex
\section{Introduction}
\IEEEPARstart{A}{mong} the inverse problems in imaging, denoising is without doubt the most extensively studied \cite{elad_survey}. In its simplest formulation, an image $x \in \mathbb{R}^d$ is perturbed by an additive white Gaussian noise (AWGN) $w$ of variance $\sigma^2$. Denoising then consists in processing the resulting noisy image $y = x + w$ in order to remove the noise component $w$ and recovering the original signal $x$.

Over the years, a rich variety of strategies, tools and theories have emerged to address this issue at the intersection of statistics, signal processing, optimization and functional analysis. But this field has been recently immensely influenced by the development of machine learning techniques and deep neural networks. Viewing denoising as a simple regression problem, this task ultimately amounts to learning a network to match the corrupted image to its source. In practice, a training phase is necessary beforehand, during which the network is optimized by stochastic gradient descent on an external dataset consisting of clean/noisy image pairs. The power of deep-learning lies in its tremendous generalization capabilities allowing it to be just as effective beyond its training set. This approach has revolutionized denoising, as well as many inverse problems in computer vision. Numerous supervised neural networks have been proposed since then for image denoising  \cite{dncnn, ffdnet, LIDIA, drunet, restormer, scunet, swinir, red30, tnrd, mwcnn, nlrn, n3net, lefkimmiatis2017non, random_shuffle_transformer, stochastic_window_transformer}, leading to state-of-the-art performance. Nowadays, the very best methods exploit the transformer architecture that is able to extract non-local features \cite{restormer, scunet, swinir, random_shuffle_transformer, stochastic_window_transformer, kong2023efficient, bai2021learning}.

\input{speed}

However, these supervised methods, in addition to being cumbersome due to the computationally demanding optimization phase, suffer from their high sensitivity to the quality of the training set. The latter must indeed provide diverse, abundant and representative examples of images; otherwise, mediocre or even totally aberrant results can be obtained afterwards. This makes their use impossible in some cases, especially when noise-free images are missing (although training on datasets  composed of noisy/noisy image pairs was studied in  \cite{noise2noise, noisier2noise, R2R, N2V, laine, N2S, zhussip2019training}). To address this issue, self-supervised learning - a machine learning technique in which only the input noisy image is used for training - with deep neural networks was investigated as an alternative strategy \cite{S2S, DIP, N2S, ZS-N2N, rethinking, N2F} but their performance is still limited when compared to their conventional counterparts \cite{BM3D, nlbayes, nlridge, WNNM, NCSR, SAIST, ksvd, PEWA, OWF, TWSC, EPLL_unsupervised}.

\IEEEpubidadjcol 

In dataset-free image denoising, BM3D \cite{BM3D} remains a reference method and is still competitive today even if it was developed fifteen years ago. Leveraging the non-local strategy, its mechanism relies on processing collaboratively groups of similar noisy patches across the image, assuming a locally sparse representation in a transform domain. Since then, a lot of methods based on patch grouping were developed \cite{nlbayes, nlridge, WNNM, SAIST, NCSR, TWSC}. %achieving comparable performance
In a recent paper \cite{nlridge}, we proposed a unified view of several non-local two-step methods, BM3D \cite{BM3D} being at the forefront. We showed how these methods can be reconciled starting from the definition of a family of estimators. Under this paradigm, we inferred a novel algorithm (NL-Ridge \cite{nlridge}) based on ridge regressions, which, despite its apparent simplicity, obtains the best performance when compared to \cite{BM3D, nlbayes}. A natural idea for improving these non-local two-step methods \cite{BM3D, nlbayes, nlridge} is to  repeat the second step iteratively, taking advantage of the availability of a supposedly better image estimate with each iteration. However, counter-intuitively, it is disappointing in practice as if these methods intrinsically peaked at the second step, with no theoretical justification.

\input{figure2}

 %We show that the denoising performance, assessed in terms of PSNR values, is also significantly improved when compared to single-image  deep-learning-based and the most competitive two-step denoisers \cite{BM3D, nlbayes, nlridge}.

In this paper, our main contributions are the following ones:
\begin{enumerate}
    \item In order to overcome the second stage limitation, we propose  to generalize the underlying parametric formulation of non-local denoisers by  a novel chaining technique. Compared to other patch-based algorithms, the proposed method does not exhibit a breakdown in reconstruction quality after two iterations. This particularity allows to iterate the process a dozen of times, improving gradually reconstruction quality in small steps \cite{delbracio2023inversion}. Therefore, the proposed algorithm named LIChI (Linear
and Iterative Combinations of patcHes for Image denoising) removes a large amount
of denoising artifacts, resulting in visually pleasant final images.
    %We show that, when iterating and by exploiting  more and more refined pilots, linear combinations of patches become unreasonably effective for single-image denoising, while being relatively fast (see Fig. \ref{speed}). Moreover, compared to the two-step NL-Ridge version \cite{nlridge}, the proposed algorithm named LIChI (Linear and Iterative Combinations of patcHes for Image denoising) removes a large amount of denoising artifacts, resulting in visually pleasant final images.
\item The LIChI algorithm achieves state-of-the-art performance among the self-supervised methods (see Fig. \ref{speed}), while being much faster (factor 8, CPU) than the
best performing WNNM \cite{WNNM} method and fully interpretable, requiring no matrix
diagonalization. This proves that leveraging only linear combinations of patches
(while WNNM is based on a low-rank and SVD decomposition approach) are
unreasonably effective for single-image denoising. Indeed, contrary to what might
sometimes be believed, deep-learning-based methods are not competitive in self-supervised denoising (see Fig. \ref{speed}).
\item We propose an original technique, grounded in the risk minimization theory and
inspired by deep learning theory, for deriving an initial pilot, and study its influence on
the final result. To our knowledge, this is the first time such a study is described. We
demonstrate that whatever the initial pilot is, similar denoising results are surprisingly
obtained.
\end{enumerate}

The remainder of the paper is organized as follows. %Section \ref{section1} presents the related work on image denoising.
In Section \ref{section2}, we describe a parametric view of non-local two-step denoisers \cite{BM3D, nlbayes, nlridge} and confirm the second stage limitation. In Section \ref{section3}, we introduce a novel chaining technique of the aforementioned denoisers. Our progressive scheme approximates the optimal parameters in a self-supervised manner when considering linear combinations of similar patches. In Section \ref{section4}, leveraging some techniques inspired from deep-learning theory, we show how to derive an initial pilot, and study its impact on the final result. Finally, in Section \ref{section5}, experimental results on popular datasets, either artificially noisy or real-world data, demonstrate that the resulting algorithm outperforms the self-supervised deep-learning-based techniques and compares favorably with the very best single-image methods \cite{TWSC, WNNM} while being much faster at execution.

%% file: speed.tex
\pgfdeclareplotmark{starBlue}{
    \node[star,star point ratio=2.25,minimum size=6pt,
          inner sep=0pt,draw=blue,solid,fill=blue] {};
}

\pgfdeclareplotmark{starRed}{
    \node[star,star point ratio=2.25,minimum size=6pt,
          inner sep=0pt,draw=red,solid,fill=red] {};
}

\pgfdeclareplotmark{starlichi}{
\node [star,star points=6, star point ratio=2.25, minimum size=6pt, inner sep=0pt,draw=color2,solid,fill=color2] {};
}

%\addtolength{\tabcolsep}{-6pt} 

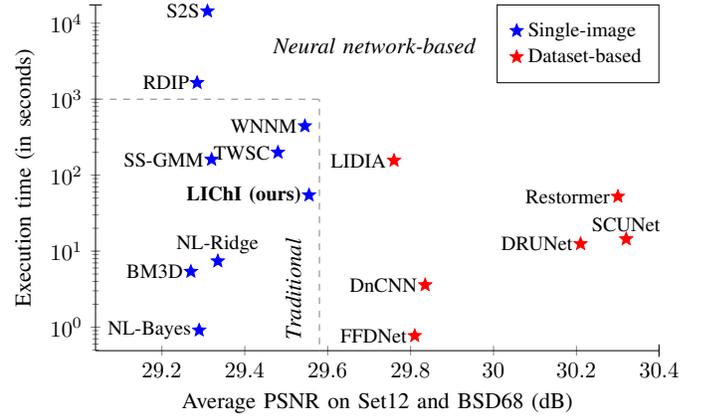
\begin{figure}[t]
\centering
\begin{tikzpicture}[scale=0.85]
\begin{axis}[
    title style={align=center},
    title={},
    cycle list name=exotic,
    ticks=both,
    %xmode=log,
    ymode=log,
    %log ticks with fixed point,
    ymin = 0.6,
    xmin = 29.04,
    xmax = 30.4,
    %y coord trafo/.code={\pgfmathparse{sqrt(#1)}},
    %y coord inv trafo/.code={\pgfmathparse{#1*#1}},
    %x coord trafo/.code={\pgfmathparse{sqrt(#1)}},
    %x coord inv trafo/.code={\pgfmathparse{#1*#1}},
    %ytick={8, 12, 16, 20, 24, 28, 32},
    %xtick={8, 12, 16, 20, 24, 28, 32},
    axis x line = bottom,
    axis y line = left,
    axis line style={-|},
    %nodes near coords = \rotatebox{45}{{\pgfmathprintnumber[fixed zerofill, precision=1]{\pgfplotspointmeta}}},
    nodes near coords align={vertical},
    every node near coord/.append style={font=\tiny, xshift=-0.5mm},
    ylabel={Execution time (in seconds)},
    xlabel={Average PSNR on Set12 and BSD68 (dB)},
    %xtick=data,
    %ymajorgrids, % for grids in gray
    %xmajorgrids,
    legend style={at={(1, 1)}, anchor=north east, legend columns=1},
    every axis legend/.append style={nodes={right}, inner sep = 0.2cm},
   %x tick label style={align=center, yshift=-0.6cm},
    %enlarge x limits=0.08,
    enlarge y limits=0.02,
    width=10.4cm,
    height=7cm,
]

    \addplot[only marks, blue,mark=starBlue, mark size=3pt ,mark options={solid}] coordinates {(29.335, 7.39)} node[midway, above] {\small \textcolor{black}{NL-Ridge}};

    \addplot[only marks, red,mark=starRed, mark size=3pt ,mark options={solid}] coordinates {(30.30, 52.39)} node[midway, left] {\small \textcolor{black}{Restormer}};

    \addplot[only marks, blue,mark=starBlue, mark size=3pt ,mark options={solid}] coordinates {(29.555, 54.63)} node[midway, left] {\small \textbf{\textcolor{black}{LIChI (ours)}}};
    
    \addplot[only marks, blue,mark=starBlue, mark size=3pt ,mark options={solid}] coordinates {(29.29, 0.911)} node[midway, left] {\small \textcolor{black}{NL-Bayes}};

    \addplot[only marks, blue,mark=starBlue, mark size=3pt ,mark options={solid}] coordinates {(29.27, 5.39)} node[midway, left] {\small \textcolor{black}{BM3D}};
    
    \addplot[only marks, blue,mark=starBlue, mark size=3pt ,mark options={solid}] coordinates {(29.545, 443.22)} node[midway, left] {\small \textcolor{black}{WNNM}};

    \addplot[only marks, red,mark=starRed, mark size=3pt ,mark options={solid}] coordinates {(30.21, 12.48)} node[midway, left] {\small \textcolor{black}{DRUNet}};

    \addplot[only marks, red,mark=starRed, mark size=3pt ,mark options={solid}] coordinates {(29.835, 3.59)} node[midway, left] {\small \textcolor{black}{DnCNN}};

    \addplot[only marks, blue,mark=starBlue, mark size=3pt ,mark options={solid}] coordinates {(29.31, 14400)} node[midway, left] {\small \textcolor{black}{S2S}};

    \addplot[only marks, blue,mark=starBlue, mark size=3pt ,mark options={solid}] coordinates {(29.285, 1645)} node[midway, left] {\small \textcolor{black}{RDIP}};

    \addplot[only marks, red,mark=starRed, mark size=3pt ,mark options={solid}] coordinates {(29.81, 0.77)} node[midway, left] {\small \textcolor{black}{FFDNet}};

    \addplot[only marks, red,mark=starRed, mark size=3pt ,mark options={solid}] coordinates {(29.76, 155.86)} node[midway, left] {\small \textcolor{black}{LIDIA}};

     \addplot[only marks, red,mark=starRed, mark size=3pt ,mark options={solid}] coordinates {(30.32, 14.43)} node[midway, above] {\small \textcolor{black}{SCUNet}};

    \addplot[only marks, blue,mark=starBlue, mark size=3pt ,mark options={solid}] coordinates {(29.32, 160.93)} node[midway, left] {\small \textcolor{black}{SS-GMM} };

    \addplot[only marks, blue,mark=starBlue, mark size=3pt ,mark options={solid}] coordinates {(29.48, 199.30)} node[midway, left] {\small \textcolor{black}{TWSC}};

    \addplot[dashed,color=gray]coordinates {(0,1000)(29.58,1000)};

    \addplot[dashed,
    color=gray]
    coordinates {
    (29.58,0.01)(29.58,1000)
    };

    \addplot[only marks,color=white,mark=*, mark size=3pt ,mark options={solid}] coordinates {(29.45, 5000)} node[midway, right] { \textcolor{black}{\textit{Neural network-based}}};

    \addplot[only marks,color=white,mark=*, mark size=3pt ,mark options={solid}] coordinates {(29.48, 3.2)} node[midway, right] {\begin{sideways} \textcolor{black}{\textit{Traditional}} \end{sideways}};

\legend{\small Single-image, \small Dataset-based}

\end{axis}
\end{tikzpicture}

\caption{The execution time on CPU for an image of size $512\times512$ v.s the average PSNR results on Set12 and BSD68 \cite{berkeley} for synthetic Gaussian noise with $\sigma=25$ of the most effective popular methods  \cite{drunet, dncnn, ffdnet, LIDIA, scunet, restormer, nlridge, nlbayes, BM3D, WNNM, EPLL_unsupervised, TWSC, rethinking, S2S}. These results are calculated based on Table \ref{resultsPSNR} and subsection \ref{section_complexity}.}
\label{speed}
\end{figure}

%% file: figure2.tex
\begin{figure*}[!t]
\centering
\begin{tabular}{ccc}
    \includegraphics[width=0.24\textwidth, trim={0cm 0.5cm 25.4cm 0},clip]{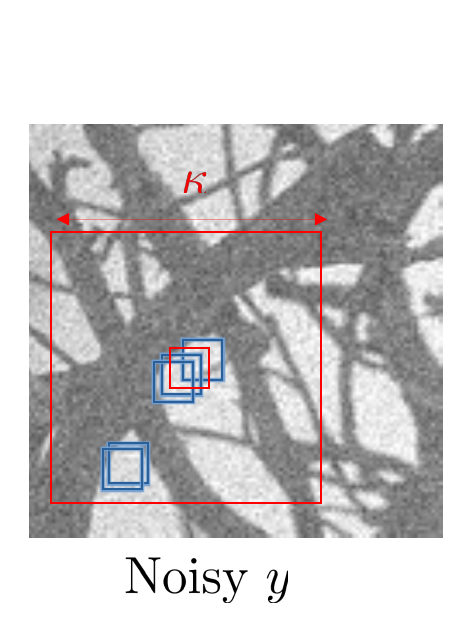}  & \includegraphics[width=0.45\textwidth, trim={7.6cm 0 8cm 0},clip]{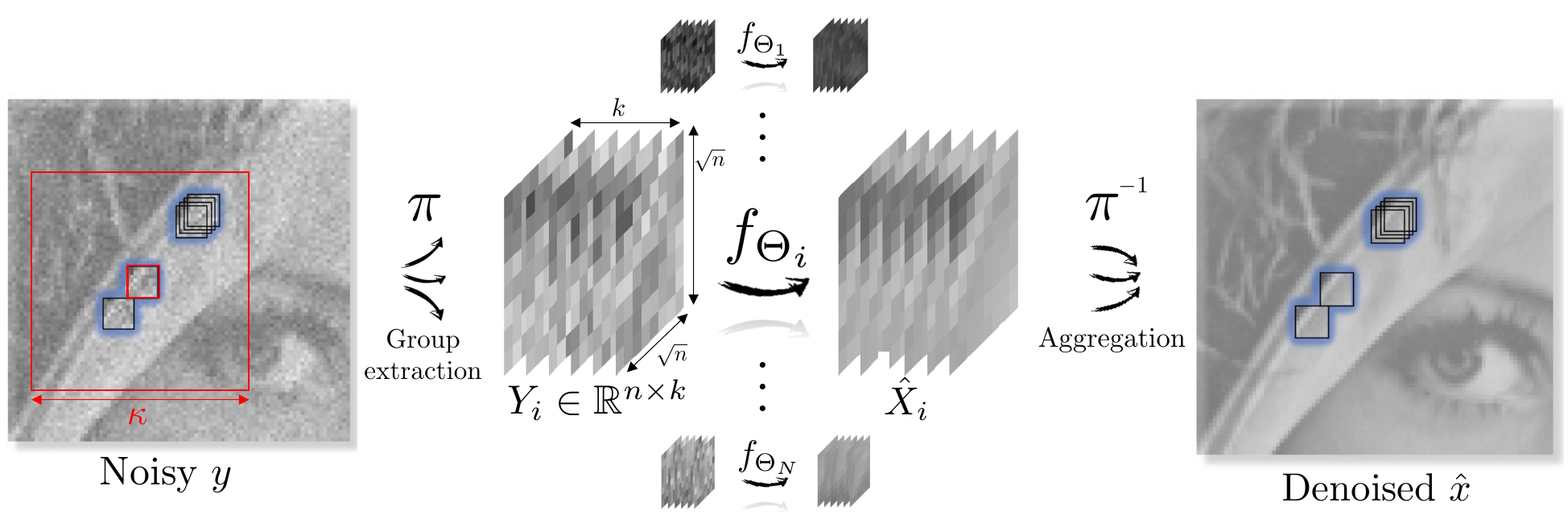}  & \includegraphics[width=0.24\textwidth, trim={25.4cm 0.5cm 0 0},clip]{BM2_nolena.pdf} \\
\end{tabular}

\renewcommand{\arraystretch}{0.8}
\begin{tabularx}{\textwidth}{CCC}
\\
   \scriptsize \it BM3D \cite{BM3D} assumes a locally sparse representation in a transform domain: & \scriptsize \it NL-Bayes \cite{nlbayes} was originally established in the Bayesian setting: & \scriptsize \it NL-Ridge \cite{nlridge} denoises each patch by a linear combination: \\
  \scalebox{0.7}{$f_{\Theta_i}(Y_i) =  P^\top (\Theta_i \odot (P Y_i Q)) Q^\top,$} & \scalebox{0.7}{$f_{\Theta_i, \beta_i}(Y_i) = \Theta_i Y_i + \beta_i \mathbf{1}_k^\top,$} & \scalebox{0.7}{$f_{\Theta_i}(Y_i) =  Y_i \Theta_i.$}\\
\scriptsize \it where $P$ and $Q$ are orthogonal matrices and $\odot$ denotes the Hadamard product. & \scriptsize \it where $\mathbf{1}_k$ denotes the $k$-dimensional all-ones vector. &    \\
\end{tabularx}
\caption{Illustration of the parametric view of several popular non-local denoisers \cite{BM3D, nlbayes, nlridge}. Examples of parameterized functions $f_{\Theta_i}$, unequivocally identifying the denoiser, are given whose parameters $\Theta_i$ are eventually selected for each group of patches by ``internal adaptation'' (see equation \eqref{risklocal1emp}).}
\label{bm}
\end{figure*}

%% file: section2.tex
\section{A parametric view of two-step non-local methods for single-image denoising}
\label{section2}

In this section, we present the framework of non-local methods and the unified formulation detailed in \cite{nlridge}, by taking an even broader view. In this line of work, we propose an extended formulation that constitutes the foundation on which we build upon in the following sections.

\subsection{A unified framework for non-local denoisers}

Popularized by BM3D \cite{BM3D}, the grouping technique (\textit{a.k.a.} block-matching) has proven to be a key ingredient in achieving state-of-the-art performance in single-image denoising \cite{nlbayes, nlridge, WNNM, SAIST, NCSR}. This technique consists in gathering noisy patches together according to their similarities in order to denoise them collaboratively. Figure \ref{bm} summarizes the whole process composed of three steps. First, groups of $k$ similar noisy square patches $\sqrt{n} \times \sqrt{n}$ are formed. For each overlapping patch $i$ taken as reference, the similarity (\textit{e.g.} $\ell_2$ sense) with its surrounding overlapping patches is computed. The $k$-nearest neighbors, including the reference patch, are then selected to form a block of similar patches encoded in a matrix $Y_i \in \mathbb{R}^{n \times k}$, where each column represents a flattened patch. Note that the number of groups of patches is strictly equal to the number $N$ of overlapping patches in the noisy input image. Subsequently, the $N$ groups are processed in parallel by applying  a local denoising function $f$. An estimate of the noise-free corresponding block of similar patches $\hat{X}_i = f(Y_i) \in \mathbb{R}^{n \times k}$ is then obtained for each group. Finally, the denoised patches are repositioned to their initial locations in the image and aggregated, or reprojected \cite{Aggreg}, as pixels may have several estimates. Generally, arithmetic (sometimes weighed) averaging is used to that end.

In this framework, the choice of the local denoising function $f$ remains an open question. Restricting it to be a member of a class of parameterized functions $(f_\Theta)$, we have proposed in \cite{nlridge} a unified framework %based on the minimization of a local statistical risk 
to properly select one candidate among the chosen class for each group of patches; Figure \ref{bm} gives the underlying parameterized functions for three different denoisers \cite{BM3D, nlbayes, nlridge}. Formally, adopting an even broader view, a non-local denoiser $\phi_{ \boldsymbol{\Theta}}$ taking as input a noisy image $y$ composed of $N$ overlapping patches of size $\sqrt{n} \times \sqrt{n}$ can itself be viewed as a high-dimensional parametric function: 
\begin{equation}
    \phi_{ \boldsymbol{\Theta}}(y) = \pi^{-1} ( F_{\boldsymbol{\Theta}} (\pi(y))) 
    \label{nonlocal}
\end{equation}
\noindent where $\pi : y \mapsto \boldsymbol{Y}$ is an operator that extracts $N$ blocks of similar patches, viewed as a third-order tensor (\textit{i.e.} three dimensional array) $\boldsymbol{Y} = \{Y_i\}_{i=1}^{N} \in \mathbb{R}^{N \times n \times k}$, $\pi^{-1}$ is its \textit{pseudo}-inverse (replacing the patches at their initial positions and aggregating them by averaging), $F_{\boldsymbol{\Theta}}$ is the function performing the denoising of all blocks of similar patches in a parallel fashion and $\boldsymbol{\Theta} = \{ \Theta_i \}_{i=1}^{N}$. More precisely, $F_{\boldsymbol{\Theta}} : \boldsymbol{Y} \in \mathbb{R}^{N \times n \times k} \mapsto \mathbb{R}^{N \times n \times k}$ is such that  $F_{\boldsymbol{\Theta}}(\boldsymbol{Y})_{i, \cdot, \cdot} = f_{\Theta_i}(\boldsymbol{Y}_{i, \cdot, \cdot}) = f_{\Theta_i}(Y_i)$ for all $i$. In other words, this function processes each group independently through $f_{\Theta_i}$ which is exclusively dedicated to the $i^{th}$ block of similar patches $Y_i$. In the following, we assume that the patch grouping operator $\pi$ is ideal and forms the patch groups solely based on the similarity of the underlying noise-free patches, and thus independently of the noise realization. This way, $\pi(y)_i$ can be identified as the $i^{th}$ noisy block of similar patches associated to the noise-free one $\pi(x)_i$. 

It is worth noting that the number of parameters of  $\phi_{ \boldsymbol{\Theta}}$ is $N$ times the number of parameters of a single local denoising function $f_{\Theta_i}$. Therefore, the number of parameters grows linearly with the number of patches. As an illustrative example, $\boldsymbol{\Theta} \in \mathbb{R}^{N \times n \times k}$ in the case of BM3D \cite{BM3D} because $\Theta_i \in \mathbb{R}^{n \times k}$ has the same size as a patch group (see Fig. \ref{bm}). This represents about a hundred million parameters to be found for a $256 \times 256$ image with standard patch and group sizes (\textit{e.g.} $n= 8 \times 8$ and $k = 16$). Fortunately, solutions do exist in practice to reduce this high number of parameters - and thus the computational burden of non-local denoisers - such as the \textit{step trick} (or sub-sampling) which is discussed in Section \ref{setting_parameters}. In addition to sub-sampling, a heuristic technique commonly used to avoid saturating memory in the case of very large images is simply to process the input image sequentially tile by tile (where a tile is a big patch of size $1,024 \times 1,024$ for example), taking care to avoid edge effects during the final assembly.

\input{pilot}

\subsection{Parameter optimization}

In \cite{nlridge}, we showed that several single-image two-step  non-local algorithms \cite{BM3D, nlbayes, nlridge} could be reconciled by adopting a local minimal risk point of view. The ultimate objective is to determine the parameters $\{\Theta_i\}_{i=1}^{N}$ by minimizing the global risk defined as:
\begin{equation}
 \mathcal{R}_{\boldsymbol{\Theta}}(x) = \mathbb{E} \|  \phi_{\boldsymbol{\Theta}}(y) - x\|_2^2, 
 \label{risk1}
\end{equation}
\noindent where $x$ is the true image and $y$ is the noisy image. %For example, in the case of additive white Gaussian noise of variance $\sigma^2$, $y \sim \mathcal{N}(x, \sigma^2 I_d)$ where $I_d$ is the identity matrix of size $d$. %The mathematical expectation is therefore solely on $y$. 
The optimal estimator is $\hat{x} =\phi_{\boldsymbol{\Theta}^\ast}(y)$ where $\boldsymbol{\Theta}^\ast$ is the minimizer of \eqref{risk1}:
\begin{equation}
 \boldsymbol{\Theta}^\ast = \arg \min_{\boldsymbol{\Theta}} \mathcal{R}_{\boldsymbol{\Theta}}(x).
 \label{risksolve1}
\end{equation}

Solving \eqref{risksolve1} directly is difficult due to the intractability of the aggregation operator $\pi^{-1}$ in \eqref{nonlocal}. Therefore, a suboptimal greedy approach is used and aims at minimizing the risk at the individual patch group level, as originally proposed in \cite{nlridge}. This allows one to decompose the problem into $N$ simpler independent subproblems:
\begin{equation}
 \Theta_i^\ast = \arg \min_{\Theta_i} \underbrace{\mathbb{E} \|  f_{\Theta_i}(Y_i) - X_i\|_F^2}_ {R_{\Theta_i}(X_i)},
 \label{risklocal1}
\end{equation}
\noindent where $Y_i = \pi(y)_i$ and $X_i = \pi(x)_i$ are the $i^{th}$ noisy and noise-free blocks of similar patches, respectively, and $\| \cdot \|_F$ denotes the Frobenius norm. The problem \eqref{risklocal1} has a closed-form solution in the case of additive white Gaussian noise as demonstrated in \cite{nlridge} for the underlying parameterized functions of \cite{BM3D, nlbayes, nlridge} illustrated in Fig. \ref{bm}.

\subsection{Principle of internal adaptation}

As the true image $x$ is not known, \eqref{risksolve1} cannot actually be solved.
However, assuming that an initial estimate $\tilde{x}$ of the denoised image (\textit{a.k.a.} pilot or oracle estimator \cite{cuisine}) is available, G. Vaksman \textit{et al.} \cite{LIDIA} proposed, in the context of deep learning, to substitute $\tilde{x}$ for $x$ in \eqref{risk1}. Formally, the idea is to consider the surrogate:
\begin{equation}
 \mathcal{R}_{\boldsymbol{\Theta}}(\tilde{x}) = \mathbb{E} \|  \phi_{\boldsymbol{\Theta}}(y) - \tilde{x}\|_2^2, 
 \label{risk1x0}
\end{equation}
\noindent where $y$ follows a distribution depending on  the noise model; in the case of additive white Gaussian noise of variance $\sigma^2$, $y \sim \mathcal{N}(\tilde{x}, \sigma^2 I_d)$.

Originally, this so-called ``internal adaptation'' technique was presented as a simple post-processing refinement to boost performance of lightweight networks already trained in a supervised manner  \cite{LIDIA, dct2net, gaintuning}. In particular, as argued in \cite{LIDIA}, the ``internal adaptation'' trick is useful if the input noisy image $y$ deviates from the general statistics of the training set. Actually, it turns out that this technique is at the core of the second stage of several state-of-the-art single-image two-step denoisers \cite{BM3D, nlbayes, nlridge} where each local risk \eqref{risklocal1} is replaced by the empirical one:
\begin{equation}
 R_{\Theta_i}(\tilde{X}_i) = \mathbb{E} \|  f_{\Theta_i}(Y_i) - \tilde{X}_i\|_F^2,
 \label{risklocal1emp}
\end{equation}
\noindent where $Y_i = \pi(y)_i$ and $\tilde{X}_i = \pi(\tilde{x})_i$.

As long as the pilot $\tilde{x}$ is not too far from the true image $x$, $\hat{x} = \phi_{\boldsymbol{\Theta}^\ast}(y)$ obtained through ``internal adaptation'' by minimizing \eqref{risk1x0} may be closer to $x$ than the pilot itself (although there is no mathematical guarantee). In practice, all two-step denoisers \cite{BM3D, nlbayes, nlridge} always observe a significant boost in performance using this technique compared to the estimator obtained after the first stage. However, counter-intuitively, repeating the process does not bring much improvement and tends on the contrary to severely degrade the image after a few iterations as attested by Figure  \ref{iterative_pilots}. Therefore, these methods stop directly after a single step of ``internal adaptation''.

In order to overcome the second stage limitation and boost performance beyond the second iteration, we introduce hereunder a generalized expression of \eqref{nonlocal}. Using a progressive optimization scheme,  our algorithm, that only performs linear combinations of patches, enables to significantly improve the denoising performance at each iteration, making it as competitive as WNNM \cite{WNNM}, the best single-image method to the best of our knowledge (see Fig. \ref{speed}).

%% file: pilot.tex
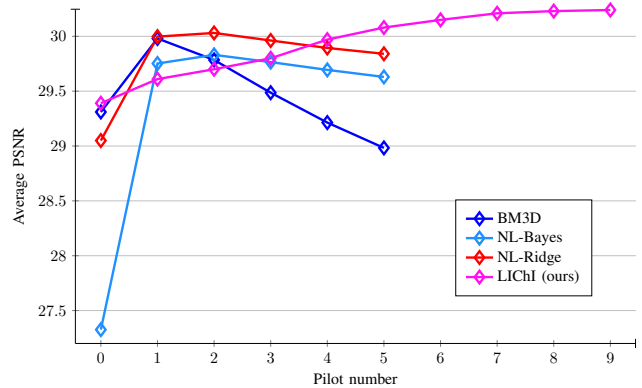
\begin{figure}[t]
\centering
\begin{tikzpicture}[scale=0.6]
\pgfplotstableread{
  29.310307919634496 27.32585708 29.049758116404217 
    29.98061595405785 29.75234886 29.997332255045574   
   29.78466664891694 29.83116468 30.030901432037354  
   29.48608137421226 29.76492882 29.96211036046346   
   29.21372458507351 29.69421109 29.894131342569988   
   28.98270836983819 29.62986271 29.840677897135418  
}\datatable

\begin{axis}[ %ybar, -> to do a bar plot
    title style={align=center},
    title={},
    cycle list name=exotic,
    ticks=both,
    ymin = 27.2,
    ymax = 30.25,
    axis x line = bottom,
    axis y line = left,
    axis line style={-|},
    ylabel={Average PSNR},
    xlabel={Pilot number},
    ymajorgrids,
    legend style={at={(0.8, 0.43)}, anchor=north, legend columns=1},
    every axis legend/.append style={nodes={right}, inner sep = 0.2cm},
    enlarge x limits=0.05,
    width=14cm,
    height=9cm,
]

    \addplot[line width=1.5pt, blue,mark=diamond, mark size=4pt , mark options={solid}] table [x expr=\coordindex, y index=0] {\datatable};
    \addplot[line width=1.5pt, color2,mark=diamond, mark size=4pt , mark options={solid}] table [x expr=\coordindex, y index=1] {\datatable};
    \addplot[line width=1.5pt, red,mark=diamond, mark size=4pt , mark options={solid}] table [x expr=\coordindex, y index=2] {\datatable};

    \addplot[line width=1.5pt, color3,mark=diamond, mark size=4pt , mark options={solid}] coordinates {(0,29.39) (1,29.61) (2,29.7)  (3,29.8)  (4,29.97)  (5,30.08)  (6,30.15)  (7,30.21)  (8,30.23)  (9,30.24)  };

\legend{BM3D, NL-Bayes, NL-Ridge, LIChI (ours)}
\end{axis}
\end{tikzpicture}

\caption{Evolution of the PSNR for pilot-based single-image denoising methods \cite{BM3D, nlbayes, nlridge} on Set12 dataset with noise level $\sigma=25$. After a PSNR jump between the first and second pilot, obtained with ``internal adaptation'', the PSNR decreases for all methods, except ours.}

\label{iterative_pilots}
\end{figure}

%% file: section3.tex
\section{LIChI: linear and iterative combinations
of patches for image denoising}
\label{section3}

In the following, we assume an additive white Gaussian noise model of variance $\sigma^2$. % for the sake of simplicity,

\subsection{A novel chaining rule for generalization}

We propose to study a class of parameterized functions that generalizes  \eqref{nonlocal}: 
\begin{equation}
    \Phi_{\{\boldsymbol{\Theta}_m\}_{m=1}^{M}}(y) =  \left[ \phi_{\boldsymbol{\Theta}_M}  \circ \ldots \circ \phi_{\boldsymbol{\Theta}_1}  \right] (y)
    \label{nonlocalstar}
\end{equation}
\noindent where $M\in \mathbb{N}^\ast$ and $\circ$ denotes the function composition operator. In other words, we consider the $M$ times iterated version of function \eqref{nonlocal}. In our approach, we focus on group denoising functions of the following straightforward form:
\begin{equation}
f_{\Theta_i}(Y_i) =  Y_i \Theta_i
\label{nlridge_group}
\end{equation}
\noindent as in \cite{nlridge} (see Fig. \ref{bm}). This choice is motivated by the fact that, despite their apparent simplicity, we proved in \cite{nlridge} that linear combinations of patches are promising for single-image denoising. Note that when fixing $\boldsymbol{\Theta}_{m} = \{I_k\}_{i=1}^{N}$ for $m \geq 2$, where $I_k$ denotes the identity matrix of size $k$, the above class of functions coincides with \eqref{nonlocal} as $f_{I_k}$ is the identity function $\operatorname{id}_{\mathbb{R}^{n \times k}}$. 

\subsection{A progressive scheme for parameter optimization}

Following the same approach as for two-step non-local denoisers, our objective is to minimize the quadratic risk:
\begin{equation}
     \{\boldsymbol{\Theta}_m^\ast\}_{m=1}^{M} = \mathop{\arg \min}\limits_{\{\boldsymbol{\Theta}_m\}_{m=1}^{M}}   \underbrace{\mathbb{E} \| \Phi_{\{\boldsymbol{\Theta}_m\}_{m=1}^{M}}(y) - x\|_2^2}_{\mathcal{R}_{\{\boldsymbol{\Theta}_m\}_{m=1}^{M}}(x)},
 \label{risk2}
\end{equation}

\noindent where $x$ is the true image assumed to be known and  $y \sim \mathcal{N}(x, \sigma^2 I)$. The optimal estimator, in the $\ell_2$ sense, is then $\hat{x} = \Phi_{\{\boldsymbol{\Theta}_m^\ast\}_{m=1}^{M}}(y)$.

Solving \eqref{risk2} is much more challenging than  minimizing \eqref{risk1} due to the repeated aggregation/extraction steps implicitly contained in expression \eqref{nonlocalstar} via  the operation $\pi \circ \pi^{-1}$. Indeed, it is worth noting that $[\pi \circ \pi^{-1}](z) \neq z$ for $z \in \mathbb{R}^{N \times n\times k}$  when patches in $z$ are not consistent (\textit{i.e.} there exists two different patch estimates for the same underlying patch). Therefore, we propose a (suboptimal) progressive approach to approximate the solution of \eqref{risk2} as follows:
\begin{equation}
\left\{
    \begin{array}{l}
       \displaystyle \boldsymbol{\Theta}^\ast_1 =  \mathop{\operatorname{argmin}}\limits_{\boldsymbol{\Theta}_1} \mathbb{E} \|  \phi_{\boldsymbol{\Theta}_1}(y) - y_1\|_2^2 \\
       \displaystyle \boldsymbol{\Theta}^\ast_2 =  \mathop{\operatorname{argmin}}\limits_{\boldsymbol{\Theta}_2} \mathbb{E} \|  [\phi_{\boldsymbol{\Theta}_2} \circ \phi_{\boldsymbol{\Theta}^\ast_1}](y) - y_2\|_2^2 \\
    \displaystyle    \quad \quad \quad \quad \quad \quad \quad \quad \quad \quad \quad \quad \vdots  \\
       \displaystyle \boldsymbol{\Theta}^\ast_M =  
       \mathop{\operatorname{argmin}}\limits_{\boldsymbol{\Theta}_M}
       \mathbb{E} \|  [\phi_{\boldsymbol{\Theta}_M} \circ \phi_{\boldsymbol{\Theta}^\ast_{M-1}} \circ \ldots \circ \phi_{\boldsymbol{\Theta}^\ast_1}](y) - y_M\|_2^2 \\
    \end{array}
\right.
 \label{scheme}
\end{equation}

\noindent where $y_m = x + \tau_m (y - x)$ with $(\tau_m)_{1 \leq m \leq M}$ a strictly decreasing sequence satisfying
$0 \leq \tau_m < 1$ and $\tau_M = 0$ (\textit{i.e.} $y_M = x$). Basically,  $\boldsymbol{\Theta}_m$ are found iteratively in a way such that composing by a new $\phi_{\boldsymbol{\Theta}_m}$ closes the gap even more with the true image $x$. Essentially, the proposed scheme amounts to solving $M$ problems of the form:
\begin{equation}
\boldsymbol{\Theta}^\ast_m =  \arg \min_{\boldsymbol{\Theta}_m} \mathbb{E} \|  \phi_{\boldsymbol{\Theta}_m} (z_{m-1}) - y_m\|_2^2,
\label{scheme_m}
\end{equation}
where $z_m = [\phi_{\boldsymbol{\Theta}^\ast_{m}} \circ \ldots \circ  \phi_{\boldsymbol{\Theta}^\ast_1}](y)$ if $m \geq 1$ and $z_0 = y$ (note that, by construction, $z_{m}$ is expected to be close to $y_m$).

\subsection{Resolution when the true image is available}

In order to solve (\ref{scheme_m}), we adopt a greedy approach by minimizing the quadratic loss at the individual patch group level as performed in (\ref{risklocal1}). The problem is then decomposed into as many independent subproblems as there are patch groups:
\begin{equation}
 \Theta_i^{m \ast} = \arg \min_{\Theta_i^m} \mathbb{E} \|  f_{\Theta_i^m}(Z_i^{m-1}) - Y_i^m\|_F^2,
 \label{risklocal2}
\end{equation}
\noindent where $Y_i^m = \pi(y_m)_i = X_i + \tau_m (Y_i - X_i)$ with $X_i= \pi (x)_i$ and $Z_i^{m-1} = \pi (z_{m-1})_i$.

In its current state, \eqref{risklocal2} cannot be solved easily as in \eqref{risklocal1} because the probability distribution of the pixels contained in $Z_i^{m-1}$ is intractable. Indeed, the repeated aggregation/extraction steps from which $Z_i^{m-1}$ is formed make obtaining its law cumbersome. However, it can be approximated by construction as a convex combination of the $i^{th}$ noisy and noise-free blocks of similar patches $Y_i = \pi(y)_i$ and $X_i = \pi(x)_i$, respectively, that is:
\begin{equation}
Z_i^{m-1} \approx X_i + t_i^{m-1} (Y_i - X_i),
\end{equation}
where $t_i^{m-1} \in (0, 1]$ is estimated for each block of similar patches and is expected to be close to $\tau_{m-1}$ when $m \geq 2$. Note that for $m=1$, this approximation is in fact exact with $t_i^{0} = 1$. Denoting $\operatorname{sd}(.)$ the operator that computes the standard deviation of the coefficients of the input random matrix, we have $\operatorname{sd}(Y_i - Z^{m-1}_i) = (1 - t_i^{m-1}) \sigma$. The parameter $t_i^{m-1}$ can therefore be estimated as follows:
\begin{equation} t_i^{m-1} = 1 - \operatorname{sd}(Y_i - Z^{m-1}_i) / \sigma.
\label{variance}
\end{equation}
\noindent Finally, conceding this small approximation, the minimizer of \eqref{risklocal2} has the following closed-form solution (see proof in Appendix):
\begin{equation}
%\begin{aligned}
%    \Theta_i^{m \ast} &= \left(1-%\frac{\tau_{m}}{t_i^{m-1}}\right)  \left( X_i^\top X_i + n (t_i^{m-1} \sigma)^2  I_k \right)^{-1} X_i^\top X_i \\
 %   &+ \frac{\tau_{m}}{t_i^{m-1}} I_k 
%\end{aligned}
    \Theta_i^{m \ast} = I_k - \left(1-\frac{\tau_{m}}{t_i^{m-1}}\right) \lambda_i^{m-1} \left( X_i^\top X_i + \lambda_i^{m-1}  I_k \right)^{-1} \,,
 \label{solvelocal2}
\end{equation}
\noindent with $\lambda_i^{m-1} = n (t_i^{m-1} \sigma)^2$.

\begin{algorithm}[t]
\caption{LIChI: Linear and Iterative Combinations of patcHes for Image denoising}
\begin{algorithmic}
\Require Noisy image $y$, initial pilot $\tilde{x}_1$, noise level $\sigma$, group size $k$, patch size $\sqrt{n}$, number of iterations $M$,
sequence $(\tau_i)_{1 \leq i \leq M}$.
\Ensure Denoised image $\hat{x}$.  
\State $z_0 = y$
\For{$m=1$ to $M$}
\For{each $\sqrt{n} \times \sqrt{n}$ patch in $z_{m-1}$ indexed by $i$} 
 \State   Find its $k$ most similar patches in $z_{m-1}$  to form \State the block of similar patches $Z^{m-1}_i$. 
 \State  Form $\tilde{X}^m_i$ and $Y^m_i$ with the corresponding  patches \State in $\tilde{x}_m$ and $y$, respectively. 
 \State  $\displaystyle  t_i^{m-1} = 1 - \operatorname{sd}(Y^m_i - Z^{m-1}_i) / \sigma$
 \State $\displaystyle \lambda_i^{m-1} = n (t_i^{m-1} \sigma)^2 $

 \State $\displaystyle \Xi_i^m = I_k - \lambda_i^{m-1} \left( \tilde{X}^{m \top}_i \tilde{X}^m_i + \lambda_i^{m-1}  I_k \right)^{-1}$

\State $\displaystyle \tilde{X}^{m+1}_i = Z^{m-1}_i \Xi_i^{m}$

\State
 $\displaystyle \Theta_i^{m} =  (1-\frac{\tau_{m}}{t_i^{m-1}})  \Xi_i^{m} + \frac{\tau_{m}}{t_i^{m-1}} I_k$   
 \State  $\displaystyle  Z^{m}_i = Z^{m-1}_i \Theta_i^{m}$ 
\EndFor 
\State Reposition and aggregate patches of each patch group \State $Z^{m}_i$ and $\tilde{X}^{m+1}_i$ to form $z_m$ and updated pilot $\tilde{x}_{m+1}$.
\EndFor 
\State \Return $z_M$
\end{algorithmic}
\label{algo1}
\end{algorithm}

\subsection{Use of multiple cost-efficient pilots to guide estimation}

%\begin{figure*}[!t]
%\centering
%\includegraphics[scale=0.5]{BM3.pdf} 
%\caption{Illustration of the proposed scheme based on the use of $M=3$ pilots for optimization.}
%\label{progressive}
%\end{figure*}

Solving the initial objective (\ref{risk2}) is impossible in practice, whatever the scheme of optimization adopted, as the true image $x$ is missing. In Section \ref{section2}, we have mentioned that substituting a pilot $\tilde{x}$ for $x$, that is applying ``internal adaptation'' \cite{LIDIA}, constitutes the reference method to overcome this issue when $M=1$. Here, we propose to use $M$ different pilots $\tilde{x}_1, \ldots, \tilde{x}_M$. More precisely, pilot $\tilde{x}_m$ is dedicated to the computation of $\boldsymbol{\Theta}^\ast_{m}$ as follows:
\begin{itemize}
    \item[-] $X_i = \pi(x)_i$ is replaced by $\tilde{X}^{m}_i = \pi(\tilde{x}_{m})_i$;
    \item[-] $t_i^{m-1}$ is computed using the sample standard deviation in (\ref{variance}) where $Y_i= \pi(y)_i$ and $Z_i^{m-1} = \pi(z_{m-1})_i$ are the only realizations at our disposal;
    \item[-] $\Theta_i^{m\ast}$ is computed with (\ref{solvelocal2}).
\end{itemize}

Let us assume that, for $m\geq1$, $\boldsymbol{\Theta}^\ast_{1}, \ldots,  \boldsymbol{\Theta}^\ast_{m-1}$ have already been computed and that a pilot $\tilde{x}_{m}$ is available. Then, $\boldsymbol{\Theta}^\ast_{m}$ can be computed using $\tilde{x}_{m}$ and we propose to update it for the next step with:
\begin{equation}
    \tilde{x}_{m+1} = [\phi_{\boldsymbol{\Xi}_{m}} \circ \phi_{\boldsymbol{\Theta}^\ast_{m-1}} \circ \ldots \circ \phi_{\boldsymbol{\Theta}^\ast_1}](y) = \phi_{\boldsymbol{\Xi}_{m}}(z_{m-1}),
    \label{pilot_xi}
\end{equation}
\noindent where parameters $\boldsymbol{\Xi}_{m}$ must be found. Ideally, we want:
\begin{equation}
    \boldsymbol{\Xi}^\ast_{m} = \arg \min_{\boldsymbol{\Xi}_{m}}   \mathbb{E} \| \phi_{\boldsymbol{\Xi}_{m}}(z_{m-1})  - x\|_2^2,
\end{equation}

\noindent for which the solution is given, according to (\ref{solvelocal2}) with $\tau_{m} = 0$, by:
\begin{equation}
\boldsymbol{\Xi}^\ast_{m} = \left\{I_k - \lambda_i^{m-1} \left( X_i^\top X_i + \lambda_i^{m-1}  I_k \right)^{-1} \right\}_{i=1}^{N}\,,
\end{equation}
\noindent with $\lambda_i^{m-1} = n (t_i^{m-1} \sigma)^2$.

 Nevertheless, as $x$ is unknown, $X_i = \pi(x)_i$  is replaced by the previous pilot, that is $\tilde{X}^{m}_i = \pi(\tilde{x}_{m})_i$, and sample standard deviation is used for the computation of $t_i^{m-1}$ in (\ref{variance}). This way, provided that an initial pilot $\tilde{x}_1$ is available, all set of matrices from $\boldsymbol{\Theta}^\ast_{1}$ to $\boldsymbol{\Theta}^\ast_{M}$ can be computed iteratively with updated pilots at each step to finally get
 \begin{equation}
 \hat{x}  = \Phi_{\{\boldsymbol{\Theta}_m^\ast\}_{m=1}^{M}}(y) = z_M= \tilde{x}_M\,,
 \end{equation}
 \noindent as the final estimate for $x$. As for the choice of the initial pilot $\tilde{x}_{1}$, the reader is referred to Section \ref{section4} in this regard. Figure \ref{iterative_pilots} shows how the $M$ pilots improve with each iteration. 
 
% ; in Fig. \ref{progressive} we illustrate the proposed scheme for efficient resolution based on the use of $M$ different pilots. 

We want to emphasize that using the $M$ proposed pilots instead of a single one for each step does not increase the computational complexity. Indeed, the cost for computing the $\boldsymbol{\Xi}^\ast_{m}$ can be immediately recycled to compute the $\boldsymbol{\Theta}^\ast_{m}$ at the same time, by noticing how close they are in expression. The whole procedure is summarized in Algorithm \ref{algo1} where patch grouping is performed on $z_m$ at step $m$.
%\begin{equation}
%    \boldsymbol{\Theta}^\ast_{m} = \left\{ (1-\frac{\tau_{m}}{t_i^{m-1}})  \Xi_i^{m \ast} + \frac{\tau_{m}}{t_i^{m-1}} I_k \right\}_{i=1}^{N}\,.
%\end{equation}

\subsection{Weighted average reprojection}
At step $m$, after processing all groups of similar patches thanks to functions $F_{\boldsymbol{\Theta}_m}$ and $F_{\boldsymbol{\Xi}_m}$, all processed patches are repositionned at their right location in the image 
and aggregated by averaging via $\pi^{-1}$ operator. If arithmetic averaging is possible for the aggregation of the pixels belonging to the same position in the image, a weighted-average reprojection is recommended in \cite{Aggreg}. As suggested in \cite{Aggreg}, each pixel belonging to column $j$ of $Z_i^{m}$ (\textit{i.e.} belonging to the $j^{th}$ patch of $Z_i^{m}$) is assigned a weight inversely proportional to the squared $\ell_2$ norm of the combination weights calculated for its processing, that is proportional to $1/\| \Theta_i^{m} e_j \|_2^2$ or $1/\| \Xi_i^{m} e_j \|_2^2$, depending on the combinations weights used, where $e_j$ is the $j^{th}$ canonical basis vector of $\mathbb{R}^k$. Those weights are such that the sum of all weights associated to a same pixel equals one.

%% file: section4.tex
\section{Building an initial pilot} 
\label{section4}

In Algorithm \ref{algo1}, an initial pilot $\tilde{x}_1$ is necessary to start. If, in theory, any denoiser can be used to that end, we show in this section how to build one of the form $\tilde{x}_1 = \phi_{\boldsymbol{\Theta}}(y)$ where linear combinations of patches are once again leveraged for local denoising (\ref{nlridge_group}). The denoisers that we consider in this section are then described by Algorithm \ref{algo2}, all differing in the estimation of the parameters $\boldsymbol{\Theta} = \{ \Theta_i \}_{i=1}^{N}$ corresponding to the combination weights. At the end, most of them yield the same denoising performance.

\begin{algorithm}[t]
\caption{Pilot computation}
\begin{algorithmic}
\Require Noisy image $y$, noise level $\sigma$,   group size $k$, patch size $\sqrt{n}$.
\Ensure Pilot estimation $\tilde{x}$.
\For{each $\sqrt{n} \times \sqrt{n}$ patch in $y$ indexed by $i$} 
\State  Find its $k$ most similar patches in $y$ to form  the block \State of similar patches $Y_i$. 
\State Compute combination weights $\Theta_i$ with (\ref{theta_sure}), (\ref{theta_nr2n}), (\ref{theta_avg})  \State or (\ref{theta_nap}). 
\State $\tilde{X}_i = Y_i \Theta_i$.
\EndFor 
\State Reposition and aggregate patches of each patch group $\tilde{X}_i$ to form the pilot image $\tilde{x}$.
\State \Return $\tilde{x}$
\end{algorithmic}
\label{algo2}
\end{algorithm}

\subsection{Stein's unbiased risk estimate (SURE)}

Considering the same risk minimization problem as (\ref{risk1}) for the optimization of $\boldsymbol{\Theta} = \{ \Theta_i \}_{i=1}^{N}$ brings us back to the study of the $N$ independent subproblems of the form (\ref{risklocal1}). However, this time we aim to minimize each local risk by getting rid of any surrogate for the ground truth blocks of similar patches $X_i$. Stein's unbiased risk estimate (SURE) is probably the most traditional choice as it only depends on the noisy image. Indeed, this popular estimate in image denoising \cite{surelet, surenlmeans, sureGM} provides an approximation of the risk $R_{\Theta_i}(X_i)$ that solely depends on the observation $Y_i$. In the case of linear combinations of patches (\ref{nlridge_group}), the computation of SURE yields (see \cite{nlridge}):
\begin{equation}
\operatorname{SURE}_{\Theta_i}(Y_i) = - kn\sigma^2 + \| Y_i \Theta_i - Y_i \|_F^2 + 2n\sigma^2  \operatorname{tr}(\Theta_i),
\end{equation}
\noindent where $\operatorname{tr}(.)$ denotes the trace operator. Substituting this estimate for the risk $R_{\Theta_i}(X_i)$ and minimizing $\operatorname{SURE}_{\Theta_i}(Y_i)$ with regards to $\Theta_i$, we get:

\begin{equation}
\Theta_i^{\text{SURE}} = I_k - n\sigma^2 \left(Y_i^\top Y_i\right)^{-1}\,.
\label{theta_sure}
\end{equation}

 \noindent Note that $\Theta_i^{\text{SURE}}$ is close to the parameters $\Theta_i^\ast$ minimizing the risk as long as the variance of
SURE is low. A rule of thumbs used in \cite{surelet} states that the number of parameters must not be ``too large'' compared to the number of data in order for the variance of SURE to remain small. In our case, this suggests that $n > k$. In other words, a small amount of large patches are necessary for applying this technique. Finally, a possible pilot for $x$ is $\tilde{x}_1 = \phi_{ \boldsymbol{\Theta}^{\text{SURE}}}(y)$ with $\boldsymbol{\Theta}^{\text{SURE}} = \{ \Theta^{ \text{SURE}}_{i} \}_{i=1}^N$.
 
\subsection{Noisier2Noise}
\label{section_n2n}

Although somewhat counter-intuitive, N. Moran \textit{et al.} \cite{noisier2noise} showed that training a dataset-based neural network to recover the original noisy image from a noisier version (synthetically generated by adding extra noise) constitutes an efficient strategy to learn denoising weights without access to any clean training examples. Transposing this idea to single-image denoising, this amounts in our case to considering the minimization of the following risk:
\begin{equation}
 \mathcal{R}_{\boldsymbol{\Theta}}(y) = \mathbb{E} \|  \phi_{\boldsymbol{\Theta}}(z) - y\|_2^2, 
 \label{riskNr2N}
\end{equation}
\noindent where $y$ is the only noisy observation at our disposal and $z$ is a noisier random vector; in the case of additive white Gaussian noise of variance $(\alpha \sigma)^2$, where $\alpha>0$ is an hyperparameter controlling the amount of extra noise, $z \sim \mathcal{N}(y, (\alpha \sigma)^2I)$. Formally, minimizing \eqref{riskNr2N} is no more difficult than minimizing \eqref{risk1} and the same greedy approximation used in \eqref{risklocal1} can be applied to solve the $N$ independent local subproblems:
\begin{equation}
\hat{\Theta}_{\alpha, i} = \arg \min_{\Theta_i} \; \mathbb{E} \|  f_{\Theta_i}(Z_i) - Y_i\|_F^2,
\label{risklocalNr2N}
\end{equation}
\noindent where $Z_i = \pi(z)_i$ and $Y_i = \pi(y)_i$. As showed in \cite{nlridge}, the problem (\ref{risklocalNr2N}) amounts to solving a multivariate ridge regression for which the closed-form solution can be found:
\begin{equation}
\hat{\Theta}_{\alpha, i} = I_k - n(\alpha\sigma)^2 \left(Y_i^\top Y_i + n(\alpha\sigma)^2 I_k\right)^{-1}\,. 
\label{risklocalNr2Nsolve}
\end{equation}

\noindent To get an estimate of the noise-free image $x$, Noisier2Noise \cite{noisier2noise} suggests to compute:
\begin{equation}
\mathbb{E}(x | z) \approx \frac{(1+\alpha^2) \phi_{\hat{\boldsymbol{\Theta}}_\alpha}(z) - z }{\alpha^2} \,,
\end{equation}
where $\hat{\boldsymbol{\Theta}}_\alpha$ is the minimizer of (\ref{riskNr2N}) approximated by $\{ \hat{\Theta}_{\alpha, i} \}_{i=1}^N$. In Appendix, we show that this quantity is equal to $\phi_{\boldsymbol{\Theta}_\alpha^{\text{Nr2N}}}(y)$ on average with $\boldsymbol{\Theta}_\alpha^{\text{Nr2N}} = \{ \Theta^{ \text{Nr2N}}_{\alpha, i} \}_{i=1}^N$ where:
\begin{equation}
\Theta^{\text{Nr2N}}_{\alpha, i} = I_k - n(1+\alpha^2)\sigma^2 \left(Y_i^\top Y_i + n(\alpha\sigma)^2 I_k\right)^{-1}\,.
\label{theta_nr2n}
\end{equation}

The choice of the hyperparameter $\alpha$ remains an open question. N. Moran \textit{et al.} \cite{noisier2noise} recommend to set $\alpha=0.5$ to handle a variety of noise levels in their experiments. Interestingly, for $\alpha \rightarrow 0$, parameters $\boldsymbol{\Theta}_\alpha^{\text{Nr2N}}$ converge to $\boldsymbol{\Theta}^{\text{SURE}}$; a practical advantage of  $\boldsymbol{\Theta}_\alpha^{\text{Nr2N}}$  over $\boldsymbol{\Theta}^{\text{SURE}}$ is that the matrices $Y_i^\top Y_i + n (\alpha \sigma)^2 I_k$ in (\ref{theta_nr2n}) are symmetric positive-definite and therefore invertible, contrary to $Y_i^\top Y_i$ in (\ref{theta_sure}) which is only positive semi-definite and positive-definite almost surely in the case of ideal additive white Gaussian noise when $n \geq k$.  For real-world noisy images, estimation through combination weights  $\boldsymbol{\Theta}_\alpha^{\text{Nr2N}}$ is recommended over $\boldsymbol{\Theta}^{\text{SURE}}$ as, in some cases, matrices $Y_i^\top Y_i$ may not be invertible. By the way, the combination weights (\ref{theta_nr2n}) can be efficiently computed based on the Cholesky factorization \cite{cholesky}.

It is worth noting that the same weight expressions as (\ref{theta_nr2n}) can also be obtained within the Recorrupted-to-Recorrupted paradigm \cite{R2R}, which was originally applied in a dataset-based deep learning context, providing an unbiased estimate of a different type of risk which is close to (\ref{risklocal1}).

\subsection{Two additional extreme pilots}

In the case where the noisy patches within a group $Y_i$ are originally strictly identical (perfect patch group), the optimal weights %, under the constraint that $\Theta_i$ is a left stochastic matrix (each column summing to $1$), 
are the ones computing an arithmetic averaging:
\begin{equation}
    \Theta_i^{\text{AVG}} =  \mathbf{1}_k \mathbf{1}_k^\top / k,
    \label{theta_avg}
\end{equation}
\noindent where $\mathbf{1}_k$ denotes the $k$-dimensional all-ones vector. Under the optimistic assumption that each patch group formed is perfect, the pilot $\tilde{x}_1 = \phi_{ \boldsymbol{\Theta^{\text{AVG}}}}(y)$ with $\Theta^{\text{AVG}} = \{ \Theta_i^{\text{AVG}} \}_{i=1}^{N}$ is then optimal. 

On the contrary, when the patch groups formed are highly dissimilar, collaborative denoising cannot be beneficial and the resulting ``do-nothing'' weights are: \begin{equation}
    \Theta_i^{\text{Noisy}} = I_k,
\label{theta_nap}
\end{equation}
\noindent where $I_k$ is the identity matrix of size $k$. Under this pessimistic assumption, the pilot $\tilde{x}_1 = \phi_{ \boldsymbol{\Theta}^{\text{Noisy}}}(y) = y$ is optimal. This amounts to considering the original noisy image itself as an initial pilot in Algorithm \ref{algo1}.

\subsection{Comparison of the pilots}

\begin{figure}[!t]
\centering
\begin{tikzpicture}[scale=0.6]
\pgfplotstableread{
34.15 38.36 38.36 37.47 38.27    37.93158436 37.98503526 29.69997295 34.16952387      36.07882309 36.4195687  27.56583929 34.34459591  
24.60 32.70 32.71 32.52 32.51       31.90361738 31.97744131 28.47142998 24.67285124    28.4802351  29.13063987 25.97491884 24.96099854 
20.17  30.23  30.24 30.18  29.91      29.31534958 29.38004684 27.58107503 20.33430465    24.49517043 25.44663477 24.58342918 20.71543487
17.24 28.61  28.61 28.60 28.28        27.64625947 27.69104497 26.48822721 17.56195847      22.67173672 23.45738856 23.1348726  17.89843273
14.15 26.82 26.81 26.86  26.44       25.80400689 25.83084679 25.06716951 14.76373553     19.82977438 20.7727623  21.21562036 15.13934056
}\datatable

\begin{axis}[ %ybar, -> to do a bar plot
    title style={align=center},
    title={},
    cycle list name=exotic,
    ticks=both,
    axis x line = bottom,
    axis y line = left,
    axis line style={-|},
    %nodes near coords = \rotatebox{45}{{\pgfmathprintnumber[fixed zerofill, precision=1]{\pgfplotspointmeta}}},
    nodes near coords align={vertical},
    every node near coord/.append style={font=\tiny, xshift=-0.5mm},
    ylabel={Output PSNR},
    xlabel={Input PSNR},
    xtick={15, 20, 25, 30, 35},
    ymajorgrids,
    xmajorgrids,
    legend style={at={(0.65, 0.05)}, anchor=south, legend columns=4},
    every axis legend/.append style={nodes={right}, inner sep = 0.2cm},
   %x tick label style={align=center, yshift=-0.6cm},
    enlarge x limits=0.1,
    width=13cm,
    height=9cm,
]
    \addplot[line width=1pt, color1,mark=o, mark options={solid}] table [x index=0:1, y index=1] {\datatable};
    \addplot[line width=1pt, color2,mark=triangle, mark options={solid}] table [x index=0:2, y index=2] {\datatable};
    \addplot[line width=1pt, color3,mark=square, mark options={solid}] table [x index=0:3, y index=3] {\datatable};
    \addplot[line width=1pt, gray,mark=pentagon, mark options={solid}] table [x index=0:4, y index=4]     {\datatable};
    
    \addplot[line width=1pt, color1, mark=o, dashed, mark options={solid}] table [x index=0:5, y index=5] {\datatable};
    \addplot[line width=1pt, color2,mark=triangle, dashed, mark options={solid}] table [x index=0:6, y index=6] {\datatable};
    \addplot[line width=1pt, color3,mark=square, dashed, mark options={solid}] table [x index=0:7, y index=7] {\datatable};
    \addplot[line width=1pt, gray,mark=pentagon, dashed, mark options={solid}] table [x index=0:8, y index=8] {\datatable};
    
    \addplot[line width=1pt, color1,mark=o, dotted, mark options={solid}] table [x index=0:9, y index=9] {\datatable};
    \addplot[line width=1pt, color2,mark=triangle, dotted, mark options={solid}] table [x index=0:10, y index=10] {\datatable};
    \addplot[line width=1pt, color3,mark=square, dotted, mark options={solid}] table [x index=0:11, y index=11] {\datatable};
    \addplot[line width=1pt, gray,mark=pentagon, dotted, mark options={solid}] table [x index=0:12, y index=12] {\datatable};

\legend{SURE \hspace*{8pt}, Nr2N \hspace*{8pt}, AVG \hspace*{8pt}, Noisy}
\end{axis}
\end{tikzpicture}
\caption[Average PSNR results at different stages of LIChI algorithm]{Average PSNR (in dB) results on patch groups (dotted line), after aggregation (dashed line) and when taken as input for Algorithm \ref{algo1} (solid line)  for Set12 dataset depending on combination weights used and noise level. Patch and group sizes are chosen as indicated by Table \ref{optimalParams}.}
\label{curves}
\end{figure}
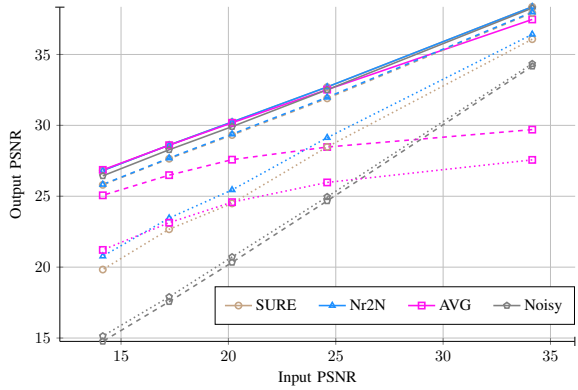

To study the performance of the proposed pilots, we examined the outputs at three different levels: \textit{i)} the individual patch group level; \textit{ii)} the global level after the aggregation stage; \textit{iii)} the output of Algorithm \ref{algo1}. Figure \ref{curves} displays the average PSNR results obtained for these three levels computed for different noise levels on Set12 dataset. Although the studied pilots have very different behaviors at the patch group level, they tend to give similar results when used as inputs for Algorithm \ref{algo1}. The ``do-nothing'' weights (\ref{theta_nap}), however, perform slightly worse than the others, especially at high noise levels, while the averaging ones (\ref{theta_avg}) are disappointing for low noise levels. As for SURE (\ref{theta_sure}) and Noisier2Noise (\ref{theta_nr2n}) weights, they give almost identical results in the end even if the Noisier2Noise weights are much more efficient on the blocks of similar patches. By the way, this highlights a non-intuitive phenomenon which was already observed in \cite{dct2net}: efficiency at the patch scale is a sufficient but not necessary condition to be efficient after the aggregation stage. This confirms that aggregation is not a basic post-processing step but plays a crucial role in image denoising. 

\begin{figure}[!t]
    \centering
    \addtolength{\tabcolsep}{-5pt}
    \renewcommand{\arraystretch}{0.5}
    \begin{tabular}{cc}
        \includegraphics[scale=0.4]{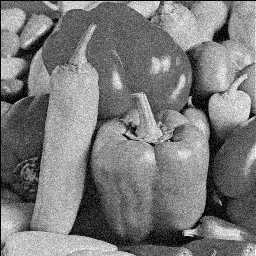}  &  \includegraphics[scale=0.47]{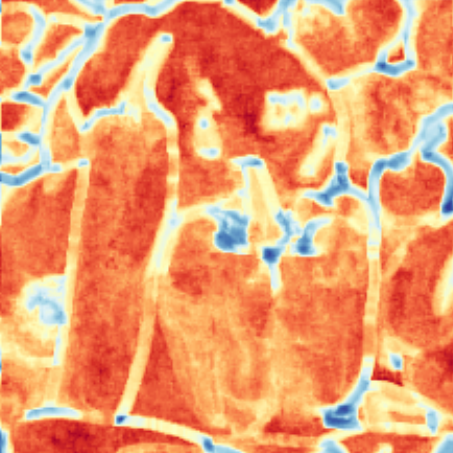}     \\
        \footnotesize  Noisy ($\sigma=15$) & \footnotesize Average / 26.83 dB   \\
        \includegraphics[scale=0.47]{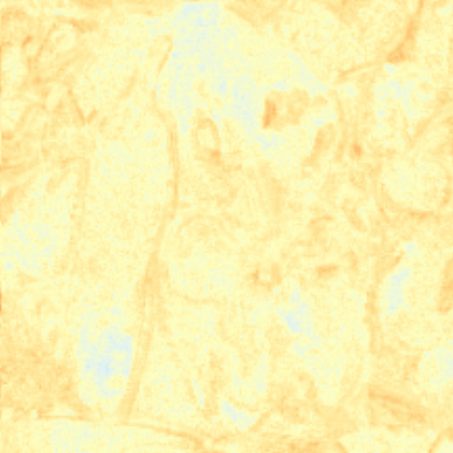} & \includegraphics[scale=0.47]{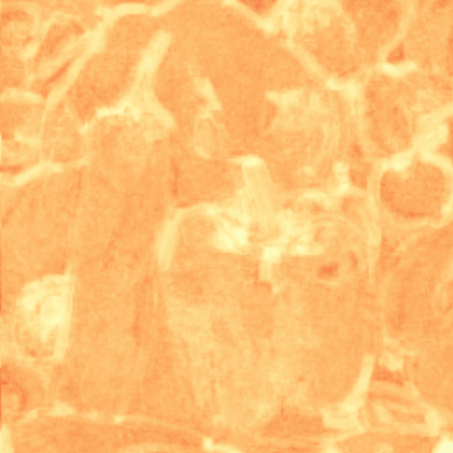}  \\
      
       \footnotesize SURE / 25.43 dB & \footnotesize  Noisier2Noise  / 28.06 dB 
    \end{tabular}  \begin{tabular}{c} \includegraphics[scale=0.6]{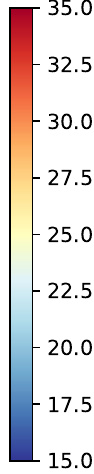} \end{tabular}
    \addtolength{\tabcolsep}{5pt} 
    
    \caption{Colormap of the PSNR (in dB) of the denoised blocks of similar patches ($n=7 \times 7$ and $k = 18$) associated with each overlapping  patch of the noisy image. The average PSNR on blocks of similar patches is also indicated.}
    \label{psnr_map}
\end{figure}

For illustration, Fig. \ref{psnr_map} provides a visual comparison of the performance of the different combination weights $\Theta_i^{\text{Name}}$ where $\text{Name} = \{\text{SURE}, \text{Nr2N}, \text{AVG}, \text{Noisy}  \}$,  depending on the location of the reference patch for intermediate noise level. Unsurprisingly, combination weights (\ref{theta_avg}) are extremely effective on the smooth parts of the image because they are theoretically optimal when applied on groups of patches being originally identical. However, when the patch groups are less homogeneous, which occurs when the reference patch is a rare patch, averaging over inherently dissimilar patches severely affects denoising. On the contrary, SURE (\ref{theta_sure}) and Noisier2Noise (\ref{theta_nr2n}) weights seem to be more versatile and less sensitive to the homogeneity of the blocks of similar patches, yielding comparable reconstruction errors regardless of the rarity of the reference patch.

\subsection{The crucial role of the aggregation stage}

\begin{figure}[!t]
    \centering
    \addtolength{\tabcolsep}{-5pt} 
    \renewcommand{\arraystretch}{0.5}
    \begin{tabular}{cc}
    
    \begin{tikzpicture}
\node[anchor=south west,inner sep=0] (image) at (0,0)
  {\includegraphics[scale=0.694925]{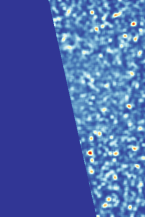}};
\begin{scope}[x={(image.south east)},y={(image.north west)}]
  \begin{scope}[shift={(0,1)},x={(1/321,0)},y={(0,-1/481)}]
    \draw[densely dashed, thick, dodgerblue] (107, 1) -> ( 214, 480);
  \end{scope}
\end{scope}
\end{tikzpicture}
    
    &
    
     \begin{tikzpicture}
\node[anchor=south west,inner sep=0] (image) at (0,0)
  {\includegraphics[scale=0.3135]{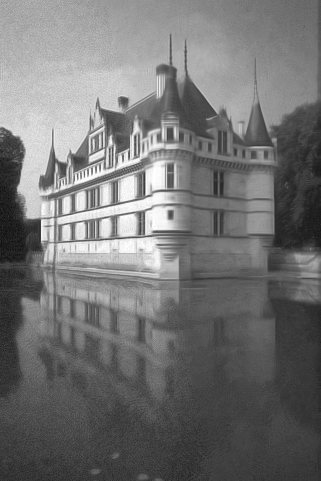}};
\begin{scope}[x={(image.south east)},y={(image.north west)}]
  \begin{scope}[shift={(0,1)},x={(1/321,0)},y={(0,-1/481)}]
    \draw[densely dashed, thick, dodgerblue] (107, 1) -> ( 214, 480);
  \end{scope}
\end{scope}
\end{tikzpicture}\\

\scriptsize Number of estimates used per pixel  & \scriptsize AVG (27.62 dB / 27.93 dB) \\

\begin{tikzpicture}
\node[anchor=south west,inner sep=0] (image) at (0,0)
  {\includegraphics[scale=0.3135]{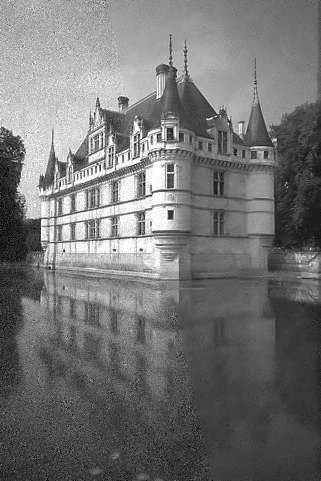}};
\begin{scope}[x={(image.south east)},y={(image.north west)}]
  \begin{scope}[shift={(0,1)},x={(1/321,0)},y={(0,-1/481)}]
    \draw[densely dashed, thick, dodgerblue] (107, 1) -> ( 214, 480);
  \end{scope}
\end{scope}
\end{tikzpicture}

&
    \begin{tikzpicture}
\node[anchor=south west,inner sep=0] (image) at (0,0)
  {\includegraphics[scale=0.3135]{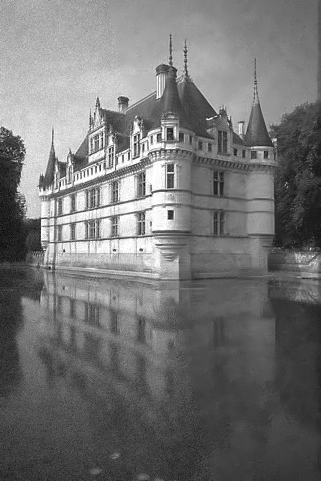}};
\begin{scope}[x={(image.south east)},y={(image.north west)}]
  \begin{scope}[shift={(0,1)},x={(1/321,0)},y={(0,-1/481)}]
    \draw[densely dashed, thick, dodgerblue] (107, 1) -> ( 214, 480);
  \end{scope}
\end{scope}
\end{tikzpicture}\\

           \scriptsize SURE   (25.64 dB / 31.52 dB) & \scriptsize Nr2N (28.79 dB / 31.83 dB) \\
    \end{tabular} \begin{tabular}{c} \includegraphics[scale=0.5]{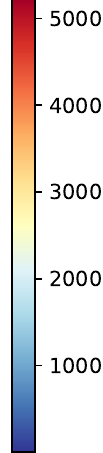} \end{tabular}
    
    \addtolength{\tabcolsep}{5pt} 
    \caption{Image denoising of \textit{Castle} image from BSD68 dataset ($\sigma=15$) by Algorithm \ref{algo2} for three different combination weights: (\ref{theta_sure}) , (\ref{theta_nr2n}) and (\ref{theta_avg}). Left: a single estimate per pixel (no aggregation), right: aggregation by averaging all estimates per pixel.} 
    \label{single-average}
\end{figure}
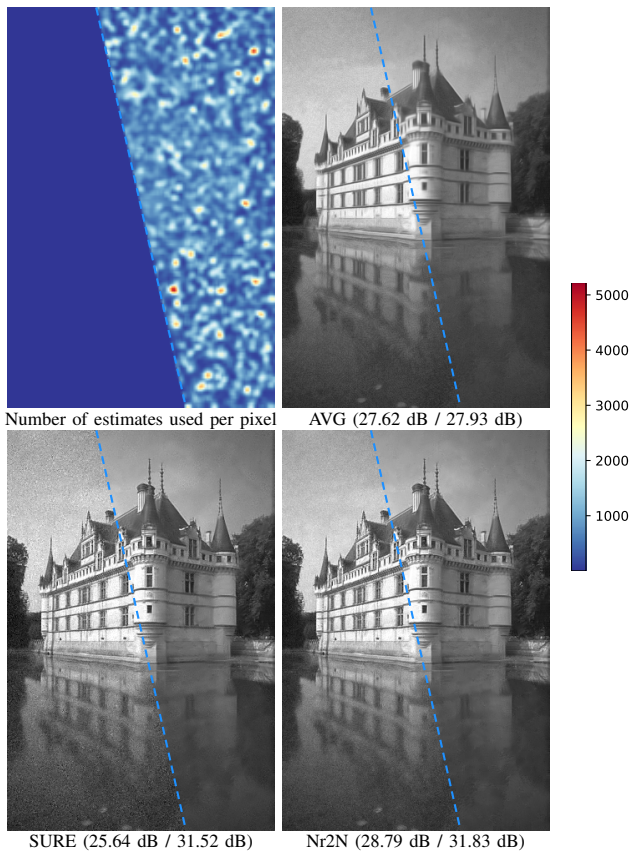

\begin{figure}
\centering
\begin{tikzpicture}[scale=0.6]
\pgfplotstableread{
 25.9758 135.4678 161.4435
 22.1474 45.1415 67.2888
27.4421 90.2347  117.6768
  24.2752 39.5961 63.8714
 117.5940 22.0008 139.5948
  137.0285 5.4276 142.4560
}\datatable

\node [align=center] at (1.90cm, -.80cm) {\scriptsize \text{SURE}};

\node [align=center] at (5.65cm, -.80cm) {\scriptsize \text{Nr2N}};

\node [align=center] at (9.6cm, -.80cm) {\scriptsize \text{AVG}};

\begin{axis}[
 ybar stacked,
    title style={align=center},
    title={},
    ticks=both,
    ytick={0, 25, 50, 75, 100, 125, 150, 175},
    axis x line = bottom,
    axis y line = left,
    axis line style={-|},
    %nodes near coords = \rotatebox{90}{{\pgfmathprintnumber[fixed zerofill, precision=2]{\pgfplotspointmeta}}},
    %nodes near coords align={vertical},
    %every node near coord/.append style={font=\small, fill=white, yshift=0.5mm},
    enlarge y limits={lower, value=0.},
    enlarge y limits={upper, value=0.},
    ylabel={MSE},
   xtick=data,
    ymin = 0,
    ymajorgrids,
    xticklabels={ 
        w/o, w/, w/o, w/, w/o, w/},
    legend style={at={(0.5, 0.95)}, anchor=north, legend columns=2},
    every axis legend/.append style={nodes={right}, inner sep = 0.2cm},
   x tick label style={align=center, yshift=-0.cm},
    enlarge x limits=0.1,
    width=13cm,
    height=7.5cm,
]
\pgfplotsinvokeforeach {0,...,1}{
    \addplot table [x expr=\coordindex, y index=#1] {\datatable};
}
\legend{Squared-bias \hspace*{8pt}, Variance} 
\end{axis}
\end{tikzpicture}
\caption{Bias-variance tradeoff between estimators (\ref{nonlocal}) and (\ref{psi}), \textit{i.e.} with (w/) and without (w/o) aggregation, for three different types of combination weights. Results obtained with noisy image \textit{Cameraman} from Set12 at noise level $\sigma=20$ with patch size $n = 9\times 9$ and group size $k=18$, estimated via Monte-Carlo simulation using $100$ different realizations of the noise.}
\label{bias-variance}
\end{figure}
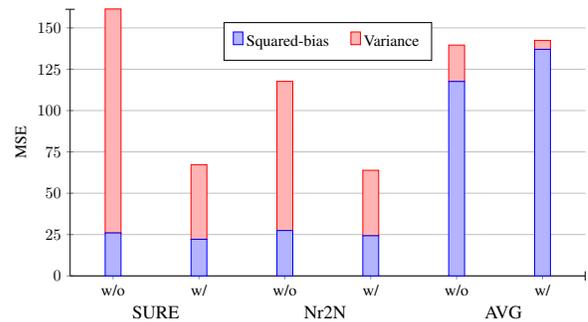

To get a better understanding of the role of the aggregation stage, let us define $\psi_{\boldsymbol{\Theta}}(y)$  the estimator that skips this operation:
\begin{equation}
    \psi_{ \boldsymbol{\Theta}}(y) = \chi ( F_{\boldsymbol{\Theta}} (\pi(y))),
    \label{psi}
\end{equation}
\noindent where operator $\chi$ replaces each patch at their initial location and selects a single estimate among those available for a given pixel. The single estimate is arbitrarily chosen at random from the most central pixels of the denoised reference patches to avoid considering poor quality estimates. In particular, when the patch size $\sqrt{n}$ is an odd number, the chosen estimates are the denoised central pixels of the reference patches. Figure \ref{single-average} illustrates the gap of performance between $\psi_{\boldsymbol{\Theta}}(y)$ and $\phi_{\boldsymbol{\Theta}}(y)$ for combination weights (\ref{theta_sure}), (\ref{theta_nr2n}) and (\ref{theta_avg}). Skipping the aggregation step results in a much poorer estimation, especially for weights (\ref{theta_sure}) and (\ref{theta_nr2n}). As a matter of fact, non-local methods have the particularity of producing a large number of estimates per noisy pixel, up to a few thousand (see Fig. \ref{single-average}), because a noisy pixel can appear in many blocks of similar patches and even several times in one. To study the benefit of exploiting those multiple estimates, a bias-variance decomposition can be leveraged:
\begin{equation}
     \underbrace{ \mathbb{E} \| \tilde{x} - x \|_2^2 / d}_{\text{MSE}} = \underbrace{\| \mathbb{E}(\tilde{x}) - x \|_2^2 / d}_{\text{squared-bias}} + \underbrace{\mathbb{E}\|   \tilde{x} - \mathbb{E}(\tilde{x}) \|_2^2 / d}_{\text{variance}} 
    \label{biasVarianceFormula}
\end{equation}
where $\tilde{x}$ is the estimator for the true image $x$. Figure \ref{bias-variance} highlights the bias–variance tradeoff for estimators $\psi_{\boldsymbol{\Theta}}(y)$ and $\phi_{\boldsymbol{\Theta}}(y)$ and combination weights (\ref{theta_sure}), (\ref{theta_nr2n}) and (\ref{theta_avg}). We can notice that the squared-bias part of the MSE in (\ref{biasVarianceFormula}) is practically unchanged  whether aggregation is applied or not. However, a  drop in variance is noticeable. This is particularly impressive for SURE estimator (\ref{theta_sure}), which significantly reduces its variance and so the MSE thanks to aggregation, closing the gap with the Noisier2Noise estimator (\ref{theta_nr2n}) as they share  almost the same squared-bias. However, for averaging estimator (\ref{theta_avg}), the variance represents a  small part in the MSE decomposition (\ref{biasVarianceFormula}) and so aggregation is not beneficial.

We can draw a parallel with a popular machine-learning ensemble meta-algorithm: bootstrap aggregating, also called bagging \cite{GB}. Bagging consists of fitting several (``weak'' in some sense) models to sampled versions of the original training dataset (bootstrap) and combining them by averaging the outputs of the models during the testing phase (aggregation). This procedure is known to improve model performance, as it decreases the variance of the model, without increasing the squared-bias. In our case, the bootstrap samples can be materialized by the numerous noisy blocks of similar patches $Y_i$ on which (weak) models $f_{\Theta_i}(.)$ are trained in an unsupervised manner. Combining pixel estimates by aggregation enables to significantly reduce the variance while keeping the squared-bias unchanged.

%% file: section5.tex
\section{Experimental results} \label{section5}

In this section, we compare the performance of the proposed single-image method, referred to as LIChI (Algorithm \ref{algo1}), with state-of-the-art methods, including related neural network-based methods \cite{dncnn, ffdnet, LIDIA, drunet, scunet, restormer, stochastic_window_transformer, DIP, N2S, S2S, N2F, ZS-N2N, rethinking} applied to standard gray images artificially corrupted with additive white Gaussian noise with zero mean and variance $\sigma^2$ and on real-world noisy images. We used the implementations provided by the authors as well as the corresponding trained weights for supervised networks. Performance of LIChI and other methods are assessed in terms of PSNR values. The code can be downloaded at: \href{https://github.com/sherbret/LIChI/}{https://github.com/sherbret/LIChI/}. %https://github.com/sherbret/LIChI/

\input{tableau}

\input{tableau_DND}
\input{figure_photo}
\input{figure_photo2}

\subsection{Setting of algorithm parameters}
\label{setting_parameters}

In all our experiments, the patch size $n$, the group size $k$ and the strictly decreasing sequence $(\tau_m)_{1 \leq m \leq M}$ in Algorithm \ref{algo1} are empirically chosen as follows: $n=6 \times 6$, $k=64$ and $\tau_m = 0.75 \times (1 - m/M)$. The number of iterations $M$ depends on the noise level $\sigma$; the higher the noise level, the more iterations of linear combinations of patches are necessary. Moreover, the optimal value of $M$ is also influenced by the quality of the initial pilot, itself depending on patch and group sizes according to Algorithm \ref{algo2}. In Table \ref{optimalParams}, we report, for each noise range, the recommended patch size $n$ and group size $k$ in algorithm \ref{algo2} with Noisier2Noise weights (with $\alpha=0.5$), as this is the most relevant choice based on the experiments in Section \ref{section4}, as well as the associated number of iterations $M$.

\begin{table}[ht]
\caption{Recommended patch size $n$ and group size $k$ for Algorithm \ref{algo2} and corresponding number of iterations $M$ in Algorithm \ref{algo1}.}
\centering
\begin{tabular}{*{4}{|c}|}
  \hline
   $\sigma$ & $n$ & $k$ & $M$ \\\hline\hline
 $\textcolor{white}{1}0< \sigma \leq 10$ & $9\times9$ & 16 & 6\\\hline
 $10 < \sigma \leq 30$ & $11\times11$ & 16 & 9\\\hline
 $30 < \sigma \leq 50$ & $13\times13$ & 16 & 11\\\hline
\end{tabular}
\label{optimalParams}
\end{table}

For the sake of computational efficiency, the search for similar patches, computed in the $\ell_2$ sense, across the image is restricted to a small local window $\kappa \times \kappa$ centered around each reference patch (in our experiments $\kappa=65$). Considering iteratively each overlapping patch of the image as reference patch is also computationally demanding, therefore only one patch out of $\delta$, both horizontally and vertically, is considered as a reference patch. The number of reference patches and thus the time spent searching for similar patches is then divided by $\delta^2$. This common technique \cite{BM3D, nlridge, WNNM} is sometimes referred in the literature as the \textit{step trick}. In our experiments, we take $\delta = 3$. Finally, to further speed up the algorithm, the search for the location of patches similar to the reference ones is only performed every third iteration because, in practice, the calculated locations vary little from one iteration to the next.

\subsection{Results on artificially noisy  and real-world noisy images}

We tested the denoising performance of our method on three well-known grayscale datasets, namely Set12, BSD68 \cite{berkeley} and Urban100 \cite{urban}, that were artificially corrupted by additive white Gaussian noise (AWGN). PSNR results are reported in Table \ref{resultsPSNR} (the missing figures are due either to the unavailability of a specific model for the noise levels concerned in the case of dataset-based methods, or to prohibitive execution times on large datasets \cite{urban} for single-image methods). For the sake of a fair comparison, algorithms are divided into two categories: single-image methods, meaning that these methods (either traditional or neural network-based) only have access to the input noisy image, and dataset-based methods that require a training phase beforehand on an external dataset. Note that only the single-image extension was considered for Noise2Self \cite{N2S} and the time-consuming ``internal adaptation'' option was not used for LIDIA \cite{LIDIA}. Results show that, although simpler conceptually, LIChI is as efficient as WNNM \cite{WNNM}, the best single-image denoiser, to the best of our knowledge.
Note that contrary to our method, WNNM is based on a low-rank and SVD decomposition approach and does not use pilot images. Such results demonstrate that linear combinations of noisy patches provide state-of-the-art performance in single-image denoising. It is worth noting that LIChI excels for low noise levels ($\sigma \leq 15$) on all datasets; however, for very high noise levels ($\sigma \geq 50$), our method seems to lose some of its effectiveness and is surpassed by \cite{WNNM, TWSC}. Finally, it is interesting to notice that, on Urban100 \cite{urban} dataset which contains abundant structural patterns and textures, some supervised neural networks \cite{dncnn, ffdnet, LIDIA} are outperformed by the best single-image methods.

Figure \ref{photo} provides  visual comparisons with several popular algorithms. LIChI compares favorably with the very best single-image methods, but also with the famous supervised network DnCNN \cite{dncnn} which is here used as a baseline for dataset-based methods (better-performing alternatives include \cite{restormer, scunet, drunet}, see Table \ref{resultsPSNR}). In particular, DnCNN \cite{dncnn}, contrary to our method, is unable to recover properly the stripes on \textit{Barbara} image (see Fig. \ref{photo}a). Moreover, the benefit of iterating linear combinations, compared to the one-pass version represented by NL-Ridge \cite{nlridge} is clearly visible. Indeed, many eye-catchy artifacts (\textit{e.g.}, see Fig. \ref{photo}b and \ref{photo}c), especially around the edges, are removed and the resulting denoised image is much more pleasant and natural.

%\subsection{Results on real-world noisy images}

In order to demonstrate the applicability of our method on real-world data, we tested the performance on Darmstadt Noise Dataset \cite{DND}. This popular dataset is composed of $50$ real-noisy photographs, for which ground-truth images are not disclosed to avoid any bias in the evaluation (denoising results can only be evaluated online at \href{https://noise.visinf.tu-darmstadt.de/}{https://noise.visinf.tu-darmstadt.de/}). The real noise on the photographs of this dataset can be modeled as a Poisson-Gaussian noise, which is further approximated with a heteroscedastic Gaussian noise whose variance is
intensity-dependent:
\begin{equation}
    y = \mathcal{N}(x, \operatorname{diag}(ax+b))
    \label{poisson_gaussian_model}
\end{equation}
\noindent where $(a, b) \in \mathbb{R}^{+}_\ast \times \mathbb{R}^{+}_\ast$ are the noise parameters and  $\operatorname{diag}$ is the vector-to-matrix diagonal operator. For each noisy image, the authors \cite{DND} calculated the adequate noise parameters $(a, b)$ based on this model and made them available to the user. Moreover, they evaluated their own network N$^3$Net \cite{n3net} on this dataset. As no training set is provided for this challenge \cite{DND}, N$^3$Net was trained  on academic image databases corrupted by synthetic heteroscedastic Gaussian noise with noise parameters $(a, b)$ similar to those previously calculated in \cite{DND}. Same training protocol was used for DnCNN \cite{dncnn}. Both DnCNN \cite{dncnn} and N$^3$Net \cite{n3net} will serve as comparison  baselines for dataset-based methods. To apply denoisers dedicated to Gaussian noise removal such as LIChI, a variance-stabilizing transformation (VST) is necessary beforehand. We used the generalized Anscombe transform \cite{Anscombe} to that end.% as recommended by \cite{DND}. 

PSNR results are reported in Table \ref{dnd_res}. It turns out that LIChI obtained the best score, surpassing BM3D \cite{BM3D} which was so far the best single-image method on this dataset, further closing the gap with dataset-based methods. This good performance is partly explained by the fact that noise in real-world photographs is generally low, which benefits our method (see Table \ref{resultsPSNR}). In terms of visual comparison (see Fig. \ref{photo2}), same conclusions can be drawn as for Gaussian noise. Indeed, our method produces much less eye-catchy artifacts than \cite{BM3D, nlridge} and is visually similar to the eye-pleasing results of a supervised neural network such as DnCNN \cite{dncnn}.

\subsection{Complexity}
\label{section_complexity}

We would like to emphasize that although LIChI is an iterative algorithm, it is relatively fast compared to its traditional and neural network-based single-image counterparts. In Fig. \ref{speed}, we have reported the running time of different denoising algorithms. It is provided for informational purposes only, since the implementation, the language used, and the machine on which the code is run strongly influence the execution time. The gap in terms of running time between dataset-based and single-image methods is explained by the fact that these latter solve optimization problems ``on the fly''.  In comparison, dataset-based methods are trained in advance and this training time, sometimes counting in days on a GPU, is not taking into account in Fig. \ref{speed}. Nevertheless, it is worth noting that traditional single-image methods \cite{nlridge, BM3D, nlbayes, WNNM, TWSC, EPLL_unsupervised} are much less computationally demanding than their state-of-the-art neural network-based counterparts \cite{rethinking, S2S}, which use time-consuming gradient descent algorithms for optimization. Finally, note that LIChI has been entirely implemented in Python with Pytorch \cite{pytorch}, enabling it to run on GPU and thus achieve even higher speeds (about 10 times faster with our implementation).

%, while traditional ones have generally closed-form solutions.

%The CPU used is a 2,3 GHz Intel Core i7 and the GPU is a GeForce RTX 2080 Ti

%% file: tableau.tex
\begin{table*}[!t]
\centering
\caption{The PSNR (dB) results of different methods applied to three datasets corrupted with synthetic white Gaussian noise. The best method among each category (dataset-based or single-image) is emphasized in red; second best is in blue.}

\resizebox{2\columnwidth}{!}{%
  \begin{NiceTabular}{|c@{\hspace{0.1cm}} c@{\hspace{0.1cm}} c c@{\hspace{0.5cm}}
c@{\hspace{0.08cm}} c@{\hspace{0.08cm}} c@{\hspace{0.08cm}} c@{\hspace{0.08cm}} c@{\hspace{0.08cm}} c@{\hspace{0.08cm}} c@{\hspace{0.08cm}} c@{\hspace{0.08cm}} c@{\hspace{0.7cm}}
c@{\hspace{0.08cm}} c@{\hspace{0.08cm}} c@{\hspace{0.08cm}} c@{\hspace{0.08cm}} c@{\hspace{0.08cm}} c@{\hspace{0.08cm}} c@{\hspace{0.08cm}} c@{\hspace{0.08cm}} c@{\hspace{0.7cm}}
c@{\hspace{0.08cm}} c@{\hspace{0.08cm}} c@{\hspace{0.08cm}} c@{\hspace{0.08cm}} c@{\hspace{0.08cm}} c@{\hspace{0.08cm}} c@{\hspace{0.08cm}} c@{\hspace{0.08cm}} c@{\hspace{0.0cm}} c|}
  \hline
 &&&  \textbf{Methods}   & \multicolumn{9}{c}{\textbf{Set12}} & \multicolumn{9}{c}{\textbf{BSD68}} & \multicolumn{9}{c}{\textbf{Urban100}} & \\\hline\hline\noalign{\vskip 0.1cm}

      &&& Noise level $\sigma$ & 5 &/&  15 &/& 25 &/&  35 &/& 50 & 5 &/&  15 &/& 25 &/&  35 &/& 50
      & 5 &/&  15 &/& 25 &/&  35 &/& 50
      & \\[0.1cm]

   \hline\noalign{\vskip 0.1cm}

\multirow{6}{*}{\begin{sideways} \scriptsize \textbf{Dataset-based}  \end{sideways}} & \multirow{6}{*}{\begin{sideways} \scriptsize \textit{Neural}  \end{sideways}} & \multirow{6}{*}{\begin{sideways} \scriptsize \textit{network}  \end{sideways}} & DnCNN \cite{dncnn} &  37.74 &/& 32.86 &/& 30.44 &/& 28.82 &/& 27.18 & 37.71 &/& 31.73 &/& 29.23 &/& 27.69  &/& 26.23 & 37.52 &/& 32.68 &/& 29.97 &/& 28.11 &/& 26.28  & \\
 &&& FFDNet \cite{ffdnet} &  38.11 &/& 32.75 &/& 30.43 &/& 28.92 &/& 27.32 & 37.80 &/& 31.63 &/& 29.19 &/& 27.73 &/& 26.29 &  38.12 &/& 32.43 &/& 29.92 &/& 28.27 &/& 26.52 & \\
 &&& LIDIA \cite{LIDIA} &  - &/& 32.85 &/& 30.41 &/& -&/&  27.19 & - &/& 31.62 &/& 29.11 &/& -&/&  26.17 & - &/& 32.80 &/& 30.12 &/& - &/& 26.51 & \\

 &&& DRUNet \cite{drunet} & \textcolor{blue}{38.64}  &/& 33.25 &/&  30.94 &/& \textcolor{blue}{29.45} &/&  27.90 & \textcolor{blue}{38.07} &/& 31.91  &/& 29.48 &/& \textcolor{blue}{28.00} &/&  26.59 & \textcolor{blue}{38.91} &/& 33.44 &/& 31.11  &/& \textcolor{blue}{29.61} &/& 27.96  & \\

 &&& Restormer \cite{restormer} & \textcolor{red}{38.70}  &/& \textcolor{black}{33.42} &/& \textcolor{black}{31.08}  &/& \textcolor{red}{29.57} &/&  \textcolor{black}{28.01} & \textcolor{red}{38.11} &/& \textcolor{black}{31.96}  &/& \textcolor{black}{29.52}  &/& \textcolor{red}{28.05} &/&  \textcolor{black}{26.62} & \textcolor{red}{39.06} &/& \textcolor{black}{33.79} &/& \textcolor{black}{31.46} &/& \textcolor{red}{30.00} &/& \textcolor{black}{28.33} & \\
 &&& SCUNet \cite{scunet} & -  &/& \textcolor{blue}{33.43} &/& \textcolor{blue}{31.09} &/& - &/&  \textcolor{blue}{28.04} & - &/& \textcolor{blue}{31.99} &/& \textcolor{blue}{29.55} &/& - &/&  \textcolor{blue}{26.67} & - &/& \textcolor{blue}{33.88} &/& \textcolor{blue}{31.58} &/& - &/& \textcolor{blue}{28.56} & \\%[0.1cm] 

 &&& Stoformer \cite{stochastic_window_transformer} & -  &/& \textcolor{red}{33.85} &/& \textcolor{red}{31.53} &/& - &/&  \textcolor{red}{28.46} & - &/& \textcolor{red}{32.57} &/& \textcolor{red}{30.06} &/& - &/&  \textcolor{red}{27.07} & - &/& \textcolor{red}{34.24} &/& \textcolor{red}{31.92} &/& - &/& \textcolor{red}{28.72} & \\%[0.1cm] 

 \hline\noalign{\vskip 0.1cm}

 \multirow{13}{*}{\begin{sideways} \scriptsize \textbf{Single-image} \end{sideways}} & \multirow{6}{*}{\begin{sideways} \scriptsize \textit{Neural} \end{sideways}} & \multirow{6}{*}{\begin{sideways} \scriptsize \textit{network} \end{sideways}} & DIP \cite{DIP}  &  -  &/&  30.12  &/& 27.54 &/& - &/& 24.67  &  - &/& 28.83 &/& 26.59 &/& - &/& 24.13  & - &/& - &/& -  &/& - &/& - &\\
 &&& N2S \cite{N2S}    &  - &/& 31.01 &/& 28.64 &/& - &/& 25.30 & - &/&  29.46 &/& 27.72 &/& - &/& 24.77  & - &/& - &/& -  &/& - &/& - &\\

  &&& N2F \cite{N2F}   &  - &/& 30.21 &/& 28.17 &/& - &/& 25.09 &  - &/& 29.82 &/& 27.79 &/& -&/&  25.05 & - &/& - &/& -  &/& - &/& - &\\

 &&& ZS-N2N \cite{ZS-N2N}   &  - &/& 30.05 &/& 27.26 &/& - &/& 23.56 &  - &/& 29.61 &/& 26.86 &/&  - &/&  23.46 & - &/& - &/& -  &/& - &/& - &\\%[0.1cm]

 &&& S2S \cite{S2S}   &  - &/& 32.07 &/& 30.02 &/& - &/& 26.49 &  - &/& 30.62 &/& 28.60 &/& -&/&  25.70 & - &/& - &/& -  &/& - &/& - &\\ 

  &&& RDIP \cite{rethinking}   &  - &/& 32.26 &/& 29.79 &/& - &/& 26.60 &  - &/& 31.21 &/& 28.78 &/& -&/&  25.81 & - &/& - &/& -  &/& - &/& - &\\

 \cdashline{2-32}\noalign{\vskip 0.1cm}

 & \multirow{7}{*}{\begin{sideways} \scriptsize \textit{Traditional} \end{sideways}} & \multirow{3}{*}{\begin{sideways} \scriptsize \textit{2-step} \end{sideways}} & BM3D \cite{BM3D}  &  38.02 &/& 32.37 &/& 29.97 &/& 28.40 &/& 26.72 &  37.55 &/& 31.07 &/& 28.57 &/& 27.08 &/& 25.62 & 38.30 &/& 32.35  &/& 29.70 &/& 27.97 &/&  25.95 &\\
 &&& NL-Bayes \cite{nlbayes} &  38.12 &/&  32.25 &/& 29.88 &/& 28.30 &/&  26.45 & 37.62 &/& 31.16 &/& 28.70 &/&  27.18 &/&  25.58 & 38.33 &/& 31.96 &/& 29.34  &/& 27.61 &/& 25.56 &\\
 
 &&& NL-Ridge \cite{nlridge}  &   38.19 &/& 32.46 &/& 30.00 &/& 28.41 &/& 26.73 &  37.67 &/& 31.20 &/& 28.67 &/& 27.14 &/& 25.67 & 38.56 &/& 32.53 &/& 29.90 &/& 28.13 &/& 26.29 &\\

\cdashline{3-32}\noalign{\vskip 0.1cm}

  &&\multirow{4}{*}{\begin{sideways} \scriptsize \textit{Multi-step} \end{sideways}} &   SS-GMM \cite{EPLL_unsupervised}  &  38.02 &/& 32.41 &/& 29.88 &/& 28.24 &/& 26.53 &  37.17 &/& 31.28 &/& 28.76 &/& 27.19 &/& 25.71 & - &/& - &/& - &/& - &/& - &\\
 
%\cdashline{3-32}\noalign{\vskip 0.1cm}

&&& TWSC \cite{TWSC}   &  38.17&/& 32.61  &/& 30.21 &/& \textcolor{blue}{28.63} &/& \textcolor{blue}{26.95} &  37.67 &/& 31.28 &/& 28.75 &/& 27.24 &/&  \textcolor{blue}{25.76} & - &/& - &/& -  &/& - &/& - &\\ 

 &&&  WNNM  \cite{WNNM} & {\textcolor{red}{38.36}} &/&  \textcolor{blue}{32.70} &/& {\textcolor{red}{30.26}} &/& {\textcolor{red}{28.69}} &/&  {\textcolor{red}{27.05}} & {\textcolor{red}{37.80}} &/& \textcolor{blue}{31.37}  &/& \textcolor{blue}{28.83} &/& \textcolor{blue}{27.30} &/& {\textcolor{red}{25.87}} &  {\textcolor{red}{38.77}} &/& \textcolor{blue}{32.97} &/& {\textcolor{red}{30.39}} &/& {\textcolor{red}{28.70}} &/& {\textcolor{red}{26.83}} &\\

 &&& \textbf{LIChI (ours)}  &  {\textcolor{red}{38.36}} &/& {\textcolor{red}{32.71}} &/& \textcolor{blue}{30.24} &/& 28.61 &/& 26.81 & {\textcolor{red}{37.80}} &/& {\textcolor{red}{31.41}} &/& {\textcolor{red}{28.87}} &/& {\textcolor{red}{27.31}} &/& 25.72 & {\textcolor{red}{38.77}} &/&  {\textcolor{red}{33.00}} &/& \textcolor{blue}{30.37} &/& \textcolor{blue}{28.59}  &/& \textcolor{blue}{26.56} &\\[0.1cm] 
 
 \hline

\end{NiceTabular}%
}

\label{resultsPSNR}
\end{table*}

%% file: tableau_DND.tex
\begin{table*}[!t]
\centering
  \caption{Denoising results on raw data on Darmstadt Noise Dataset (DND). Best method among each category  is in bold.}
%\resizebox{2\columnwidth}{!}{%
  \begin{NiceTabular}{|c|ccc ccc c |cc|}
  \hline
  &  \multicolumn{7}{c|}{\textit{Single-image}} & \multicolumn{2}{c|}{\textit{Dataset-based}} \\\hline
  \textbf{Methods} & BM3D \cite{BM3D} &  NL-Ridge \cite{nlridge} & KSVD \cite{ksvd} & PEWA \cite{PEWA} & NCSR  \cite{NCSR}  & WNNM \cite{WNNM}    &  \textbf{LIChI}   & DnCNN \cite{dncnn} & N$^3$Net \cite{n3net}  \\\hline\hline
  \textbf{PSNR (in dB)} & 47.15  &  47.01  & 46.87 & 46.72 & 47.07    & 47.05    &  \textbf{47.35}   & 47.37 & \textbf{47.56} \\\hline
\end{NiceTabular}%
%}
\label{dnd_res}
\end{table*}

%% file: figure_photo.tex
\addtolength{\tabcolsep}{-5pt} 
\begin{figure*}[!ht]
\centering
\renewcommand{\arraystretch}{0.5}
\begin{tabular}{ccc}

(a)&
\begin{tabular}{c}
\begin{tikzpicture}
\node[anchor=south west,inner sep=0] (image) at (0,0)
  {\includegraphics[width=0.24\textwidth]{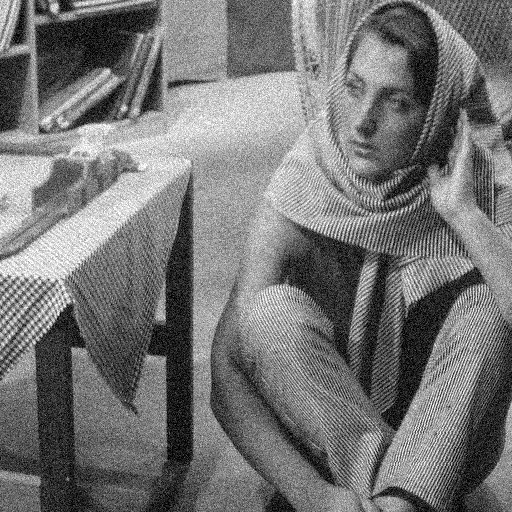}};

\node at (image.south east) [anchor=south east, inner sep=0] {\includegraphics[width=0.14\textwidth]{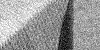}};

\begin{scope}[x={(image.south east)},y={(image.north west)}]
  \begin{scope}[shift={(0,1)},x={(1/512,0)},y={(0,-1/512)}]
  
   \draw[dodgerblue, semithick] (120, 180) rectangle (220, 230);
   
   \draw[dodgerblue, thick] (212, 362) rectangle (512, 512);

    \draw[densely dashed, thin, dodgerblue] (120, 230) -> (212, 512);
    \draw[densely dashed, thin, dodgerblue] (220, 180) -> (512, 362);
  \end{scope}
\end{scope}
\end{tikzpicture} \\
\scriptsize Noisy / 22.09 dB
\end{tabular}
& 
\begin{tabular}{ccc}
\includegraphics[width=0.225\textwidth]{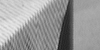} & \includegraphics[width=0.225\textwidth]{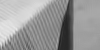} & \includegraphics[width=0.225\textwidth]{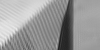} \\
\scriptsize Ground truth & \scriptsize  BM3D \cite{BM3D} / 31.72 dB & \scriptsize NL-Ridge \cite{nlridge} / 32.07 dB  \\

\includegraphics[width=0.225\textwidth]{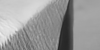} & \includegraphics[width=0.225\textwidth]{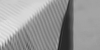} & \includegraphics[width=0.225\textwidth]{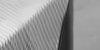} \\
\scriptsize DnCNN \cite{dncnn} / 31.06 dB  & \scriptsize  WNNM \cite{WNNM} / \textbf{32.16} dB & \scriptsize LIChI (ours)  / 32.15 dB \\
\end{tabular}

\\
& \begin{tabular}{c}\includegraphics[width=0.225\textwidth, trim=0px 20px 0px 200px, clip]{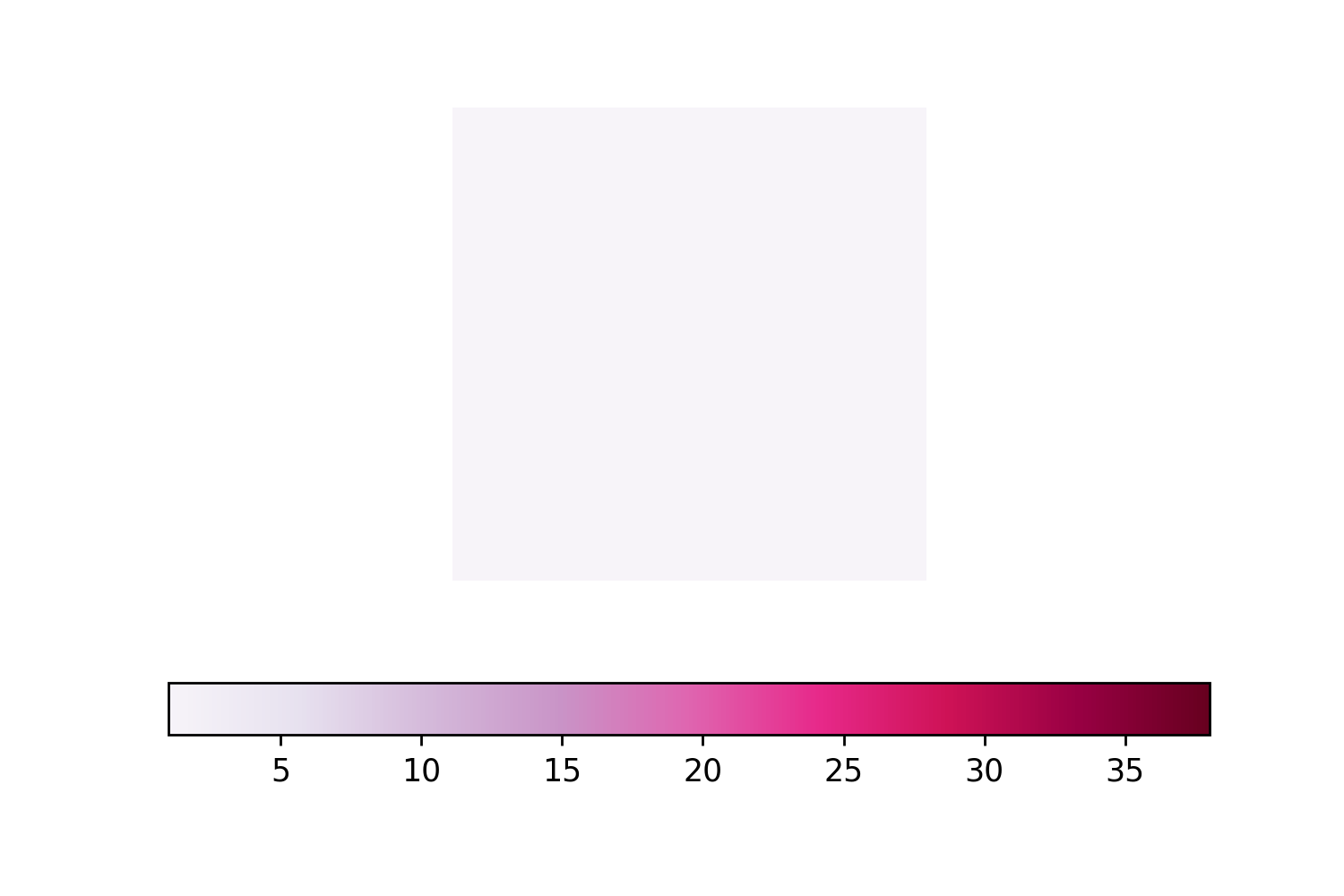} \\ \scriptsize Absolute error $|\hat{x} - x|$ \end{tabular} & 

\begin{tabular}{ccccc}
\begin{tikzpicture}
\node[anchor=south west,inner sep=0] (image) at (0,0)
{\includegraphics[width=0.135\textwidth]{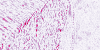}};
\begin{scope}[x={(image.south east)},y={(image.north west)}]
  \begin{scope}[shift={(0,1)},x={(1/96,0)},y={(0,-1/96)}]
   \node [color=black] at (48, 15) {\scriptsize BM3D};
  \end{scope}
\end{scope}
\end{tikzpicture}

 & 

 \begin{tikzpicture}
\node[anchor=south west,inner sep=0] (image) at (0,0)
{\includegraphics[width=0.135\textwidth]{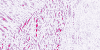}};
\begin{scope}[x={(image.south east)},y={(image.north west)}]
  \begin{scope}[shift={(0,1)},x={(1/96,0)},y={(0,-1/96)}]
   \node [color=black] at (48, 15) {\scriptsize NL-Ridge};
  \end{scope}
\end{scope}
\end{tikzpicture}

 & 

 \begin{tikzpicture}
\node[anchor=south west,inner sep=0] (image) at (0,0)
{\includegraphics[width=0.135\textwidth]{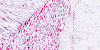}};
\begin{scope}[x={(image.south east)},y={(image.north west)}]
  \begin{scope}[shift={(0,1)},x={(1/96,0)},y={(0,-1/96)}]
   \node [color=black] at (48, 15) {\scriptsize DnCNN};
  \end{scope}
\end{scope}
\end{tikzpicture}

 & 
 \begin{tikzpicture}
\node[anchor=south west,inner sep=0] (image) at (0,0)
{\includegraphics[width=0.135\textwidth]{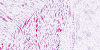}};
\begin{scope}[x={(image.south east)},y={(image.north west)}]
  \begin{scope}[shift={(0,1)},x={(1/96,0)},y={(0,-1/96)}]
   \node [color=black] at (48, 15) {\scriptsize WNNM};
  \end{scope}
\end{scope}
\end{tikzpicture}

& 

 \begin{tikzpicture}
\node[anchor=south west,inner sep=0] (image) at (0,0)
{\includegraphics[width=0.135\textwidth]{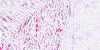}};
\begin{scope}[x={(image.south east)},y={(image.north west)}]
  \begin{scope}[shift={(0,1)},x={(1/96,0)},y={(0,-1/96)}]
   \node [color=black] at (48, 15) {\scriptsize LIChI};
  \end{scope}
\end{scope}
\end{tikzpicture}

  \\
\end{tabular}
\\
\\

(b)&
\begin{tabular}{c}
\begin{tikzpicture}
\node[anchor=south west,inner sep=0] (image) at (0,0)
% img[20:20+700, 150:150+700] i, j, m, n = 120, 180, 50, 100
  {\includegraphics[width=0.24\textwidth]{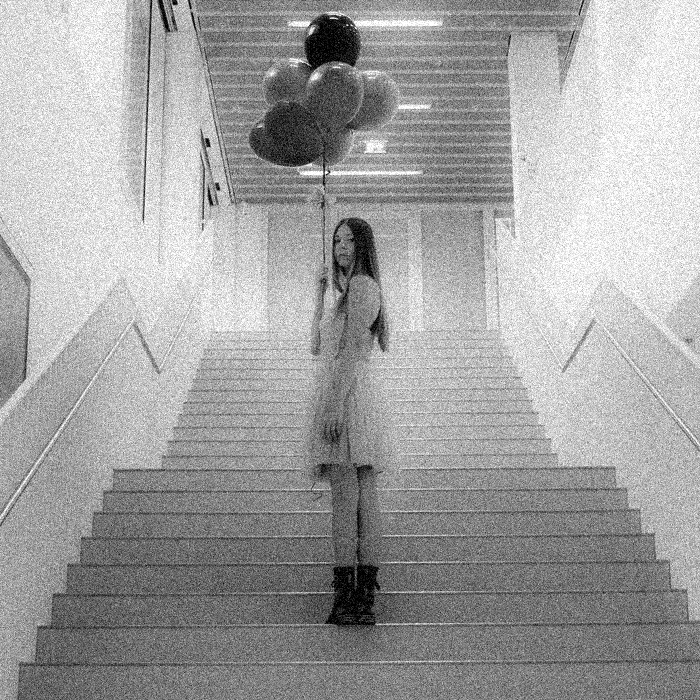}};
  
\node at (image.south east) [anchor=south east, inner sep=0] {\includegraphics[width=0.14\textwidth]{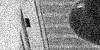}};

 \begin{scope}[x={(image.south east)},y={(image.north west)}]
  \begin{scope}[shift={(0,1)},x={(1/512,0)},y={(0,-1/512)}]
  
   \draw[dodgerblue, semithick] (131.657142857 , 87.7714285714) rectangle (204.8 , 124.342857143  );
   \draw[dodgerblue, thick] (212, 362) rectangle (512, 512);
    \draw[densely dashed, thin, dodgerblue] (131.657142857, 124.342857143 ) -> (212, 512);
    \draw[densely dashed, thin, dodgerblue] (204.8, 87.7714285714) -> (512, 362);
  \end{scope}
\end{scope}

\end{tikzpicture} \\
\scriptsize Noisy / 22.09 dB
\end{tabular}
& 
\begin{tabular}{ccc}
\includegraphics[width=0.225\textwidth]{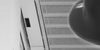} & \includegraphics[width=0.225\textwidth]{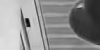} & \includegraphics[width=0.225\textwidth]{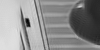} \\
\scriptsize Ground truth & \scriptsize  BM3D \cite{BM3D} / 36.76 dB & \scriptsize NL-Ridge \cite{nlridge} / 36.82 dB  \\

\includegraphics[width=0.225\textwidth]{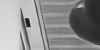} & \includegraphics[width=0.225\textwidth]{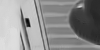} & \includegraphics[width=0.225\textwidth]{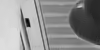} \\
\scriptsize DnCNN \cite{dncnn} / 37.06 dB  & \scriptsize  WNNM \cite{WNNM} / 36.95 dB & \scriptsize LIChI (ours) / \textbf{37.18 dB} \\
\end{tabular}

\\
& \begin{tabular}{c}\includegraphics[width=0.225\textwidth, trim=0px 20px 0px 200px, clip]{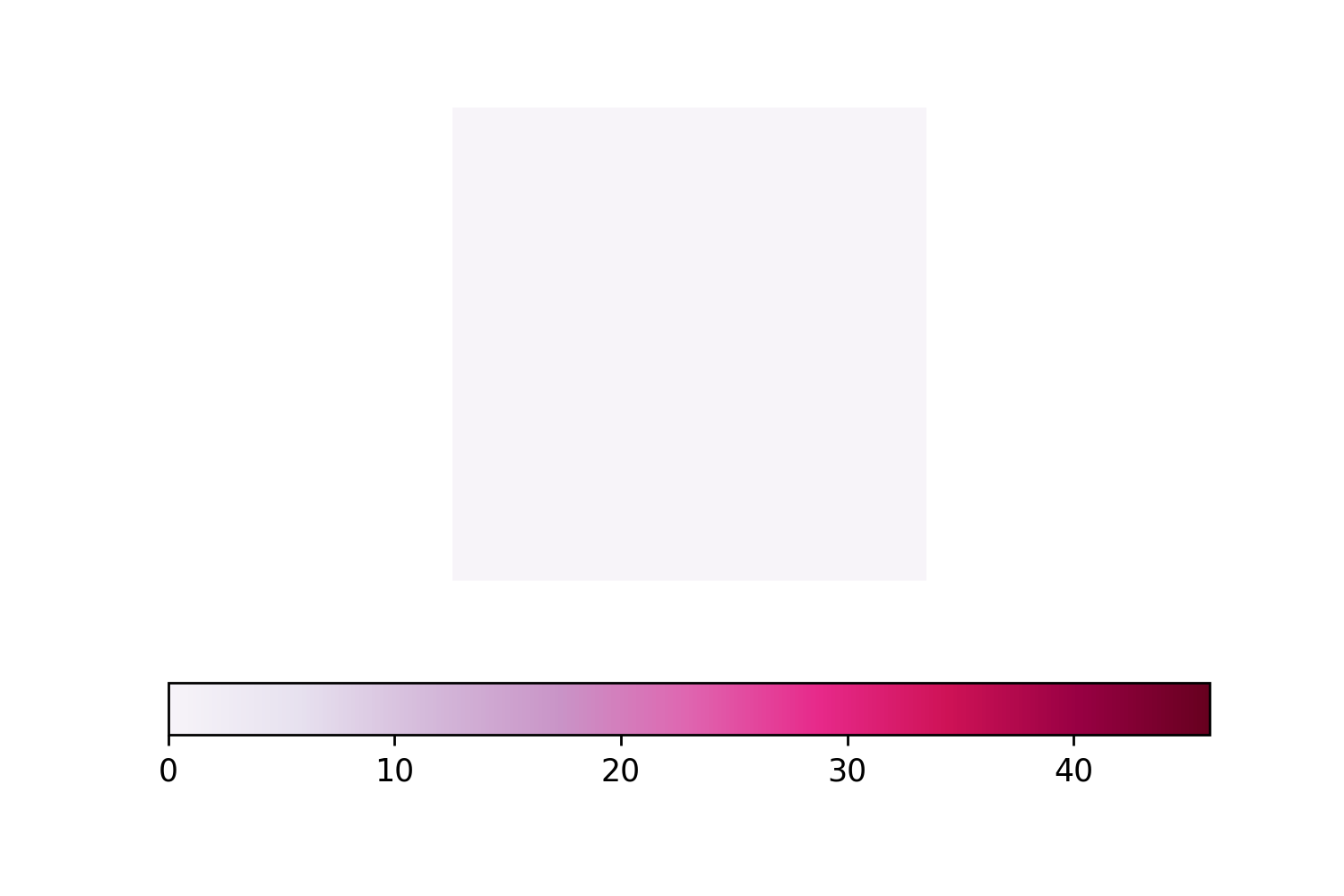} \\ \scriptsize Absolute error $|\hat{x} - x|$ \end{tabular} & 

\begin{tabular}{ccccc}
\begin{tikzpicture}
\node[anchor=south west,inner sep=0] (image) at (0,0)
{\includegraphics[width=0.135\textwidth]{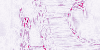}};
\begin{scope}[x={(image.south east)},y={(image.north west)}]
  \begin{scope}[shift={(0,1)},x={(1/96,0)},y={(0,-1/96)}]
   \node [color=black] at (48, 15) {\scriptsize BM3D};
  \end{scope}
\end{scope}
\end{tikzpicture}

 & 

 \begin{tikzpicture}
\node[anchor=south west,inner sep=0] (image) at (0,0)
{\includegraphics[width=0.135\textwidth]{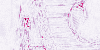}};
\begin{scope}[x={(image.south east)},y={(image.north west)}]
  \begin{scope}[shift={(0,1)},x={(1/96,0)},y={(0,-1/96)}]
   \node [color=black] at (48, 15) {\scriptsize NL-Ridge};
  \end{scope}
\end{scope}
\end{tikzpicture}

 & 

 \begin{tikzpicture}
\node[anchor=south west,inner sep=0] (image) at (0,0)
{\includegraphics[width=0.135\textwidth]{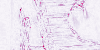}};
\begin{scope}[x={(image.south east)},y={(image.north west)}]
  \begin{scope}[shift={(0,1)},x={(1/96,0)},y={(0,-1/96)}]
   \node [color=black] at (48, 15) {\scriptsize DnCNN};
  \end{scope}
\end{scope}
\end{tikzpicture}

 & 
 \begin{tikzpicture}
\node[anchor=south west,inner sep=0] (image) at (0,0)
{\includegraphics[width=0.135\textwidth]{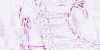}};
\begin{scope}[x={(image.south east)},y={(image.north west)}]
  \begin{scope}[shift={(0,1)},x={(1/96,0)},y={(0,-1/96)}]
   \node [color=black] at (48, 15) {\scriptsize WNNM};
  \end{scope}
\end{scope}
\end{tikzpicture}

& 

 \begin{tikzpicture}
\node[anchor=south west,inner sep=0] (image) at (0,0)
{\includegraphics[width=0.135\textwidth]{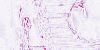}};
\begin{scope}[x={(image.south east)},y={(image.north west)}]
  \begin{scope}[shift={(0,1)},x={(1/96,0)},y={(0,-1/96)}]
   \node [color=black] at (48, 15) {\scriptsize LIChI};
  \end{scope}
\end{scope}
\end{tikzpicture}
  
\end{tabular}

\\
\\

(c)&
\begin{tabular}{c}
\begin{tikzpicture}
\node[anchor=south west,inner sep=0] (image) at (0,0)
% img[100:100+580, 200:200+580] i, j, m, n = 100, 20, 70, 140
  {\includegraphics[width=0.24\textwidth]{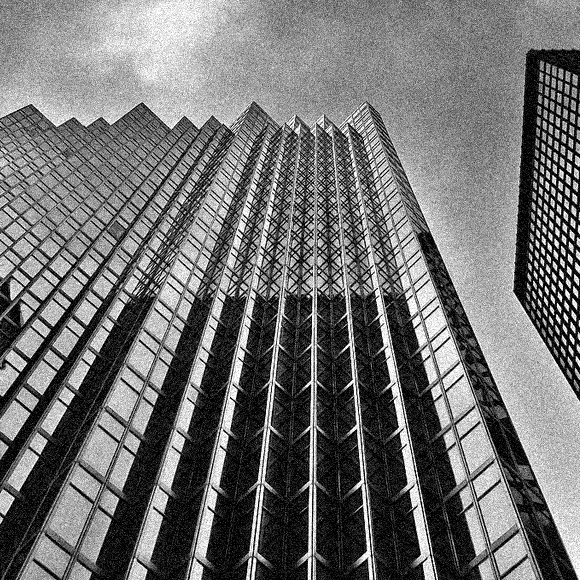}};
  
\node at (image.south east) [anchor=south east, inner sep=0] {\includegraphics[width=0.14\textwidth]{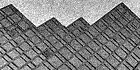}};

\begin{scope}[x={(image.south east)},y={(image.north west)}]
  \begin{scope}[shift={(0,1)},x={(1/512,0)},y={(0,-1/512)}]
  
   \draw[dodgerblue, semithick] (17.6551724138, 88.275862069 ) rectangle (141.24137931 , 150.068965517 );
   \draw[dodgerblue, thick] (212, 362) rectangle (512, 512);
    \draw[densely dashed, thin, dodgerblue] (17.6551724138, 150.068965517) -> (212, 512);
    \draw[densely dashed, thin, dodgerblue] (141.24137931, 88.275862069) -> (512, 362);
  \end{scope}
\end{scope}

\end{tikzpicture} \\
\scriptsize Noisy / 22.09 dB
\end{tabular}
& 
\begin{tabular}{ccc}
\includegraphics[width=0.225\textwidth]{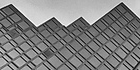} & \includegraphics[width=0.225\textwidth]{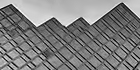} & \includegraphics[width=0.225\textwidth]{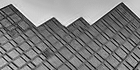} \\
\scriptsize Ground truth & \scriptsize  BM3D \cite{BM3D} / 29.50 dB & \scriptsize NL-Ridge \cite{nlridge} / 29.99 dB  \\

\includegraphics[width=0.225\textwidth]{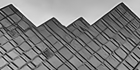} & \includegraphics[width=0.225\textwidth]{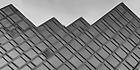} & \includegraphics[width=0.225\textwidth]{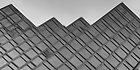} \\
\scriptsize DnCNN \cite{dncnn} / 30.06 dB  & \scriptsize  WNNM \cite{WNNM} / 30.94 dB & \scriptsize LIChI (ours) / \textbf{30.99 dB} \\
\end{tabular}

\\
& \begin{tabular}{c}\includegraphics[width=0.225\textwidth, trim=0px 20px 0px 200px, clip]{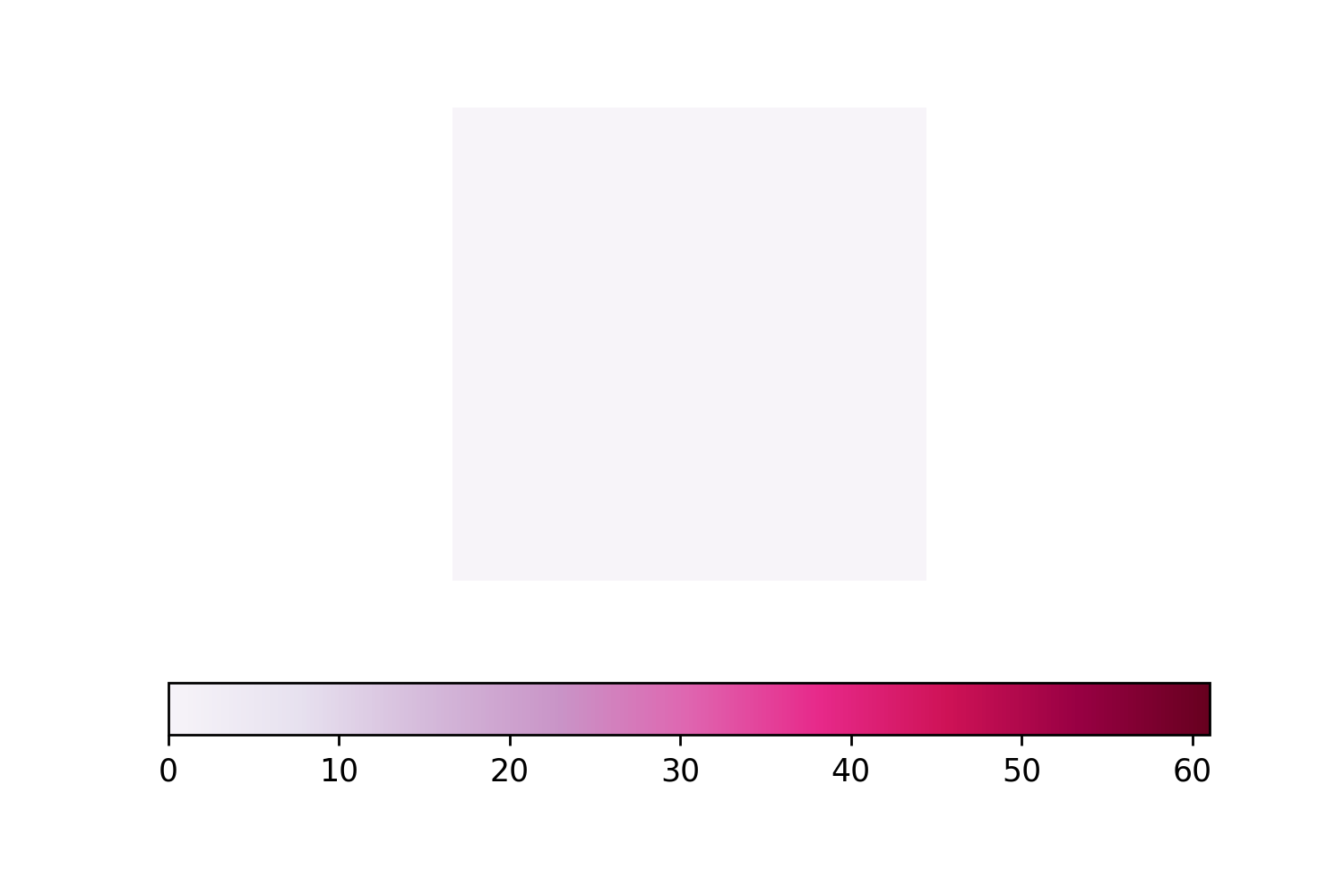} \\ \scriptsize Absolute error $|\hat{x} - x|$ \end{tabular} & 

\begin{tabular}{ccccc}
\begin{tikzpicture}
\node[anchor=south west,inner sep=0] (image) at (0,0)
{\includegraphics[width=0.135\textwidth]{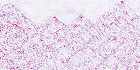}};
\begin{scope}[x={(image.south east)},y={(image.north west)}]
  \begin{scope}[shift={(0,1)},x={(1/96,0)},y={(0,-1/96)}]
   \node [color=black] at (48, 15) {\scriptsize BM3D};
  \end{scope}
\end{scope}
\end{tikzpicture}

 & 

 \begin{tikzpicture}
\node[anchor=south west,inner sep=0] (image) at (0,0)
{\includegraphics[width=0.135\textwidth]{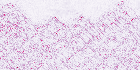}};
\begin{scope}[x={(image.south east)},y={(image.north west)}]
  \begin{scope}[shift={(0,1)},x={(1/96,0)},y={(0,-1/96)}]
   \node [color=black] at (48, 15) {\scriptsize NL-Ridge};
  \end{scope}
\end{scope}
\end{tikzpicture}

 & 

 \begin{tikzpicture}
\node[anchor=south west,inner sep=0] (image) at (0,0)
{\includegraphics[width=0.135\textwidth]{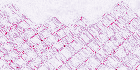}};
\begin{scope}[x={(image.south east)},y={(image.north west)}]
  \begin{scope}[shift={(0,1)},x={(1/96,0)},y={(0,-1/96)}]
   \node [color=black] at (48, 15) {\scriptsize DnCNN};
  \end{scope}
\end{scope}
\end{tikzpicture}

 & 
 \begin{tikzpicture}
\node[anchor=south west,inner sep=0] (image) at (0,0)
{\includegraphics[width=0.135\textwidth]{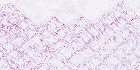}};
\begin{scope}[x={(image.south east)},y={(image.north west)}]
  \begin{scope}[shift={(0,1)},x={(1/96,0)},y={(0,-1/96)}]
   \node [color=black] at (48, 15) {\scriptsize WNNM};
  \end{scope}
\end{scope}
\end{tikzpicture}

& 

 \begin{tikzpicture}
\node[anchor=south west,inner sep=0] (image) at (0,0)
{\includegraphics[width=0.135\textwidth]{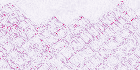}};
\begin{scope}[x={(image.south east)},y={(image.north west)}]
  \begin{scope}[shift={(0,1)},x={(1/96,0)},y={(0,-1/96)}]
   \node [color=black] at (48, 15) {\scriptsize LIChI};
  \end{scope}
\end{scope}
\end{tikzpicture}

  \\
\end{tabular}

%\footnotesize Noisy / 22.09 dB  &  \footnotesize Noisy / 22.09 dB \\

\end{tabular}
\caption{Qualitative comparison of image denoising results with synthetic white Gaussian noise ($\sigma = 20$). Zoom-in regions are indicated for each method and absolute errors to ground truth are shown for better visualization}. From top to bottom: \textit{Barbara} from Set12, \textit{Img009} from Urban100 and \textit{Img019} from Urban100.  Best viewed by zooming on a computer screen.
\label{photo}
\end{figure*}
\addtolength{\tabcolsep}{5pt}

%% file: figure_photo2.tex
\addtolength{\tabcolsep}{-5pt} 
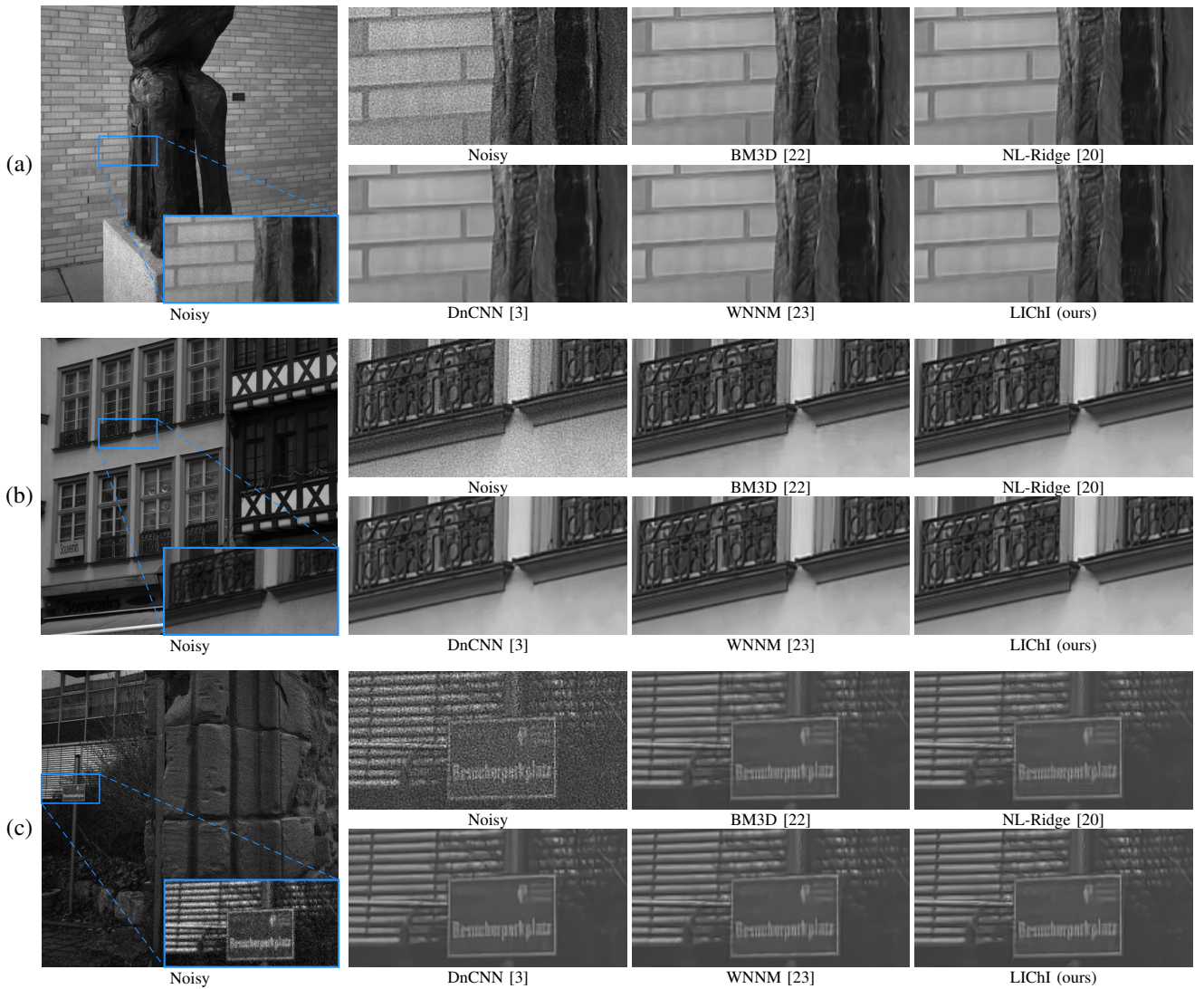
\begin{figure*}[!ht]
\centering
\renewcommand{\arraystretch}{0.5}
\begin{tabular}{ccc}

(a)&
\begin{tabular}{c}
\begin{tikzpicture}
\node[anchor=south west,inner sep=0] (image) at (0,0)
  {\includegraphics[width=0.24\textwidth]{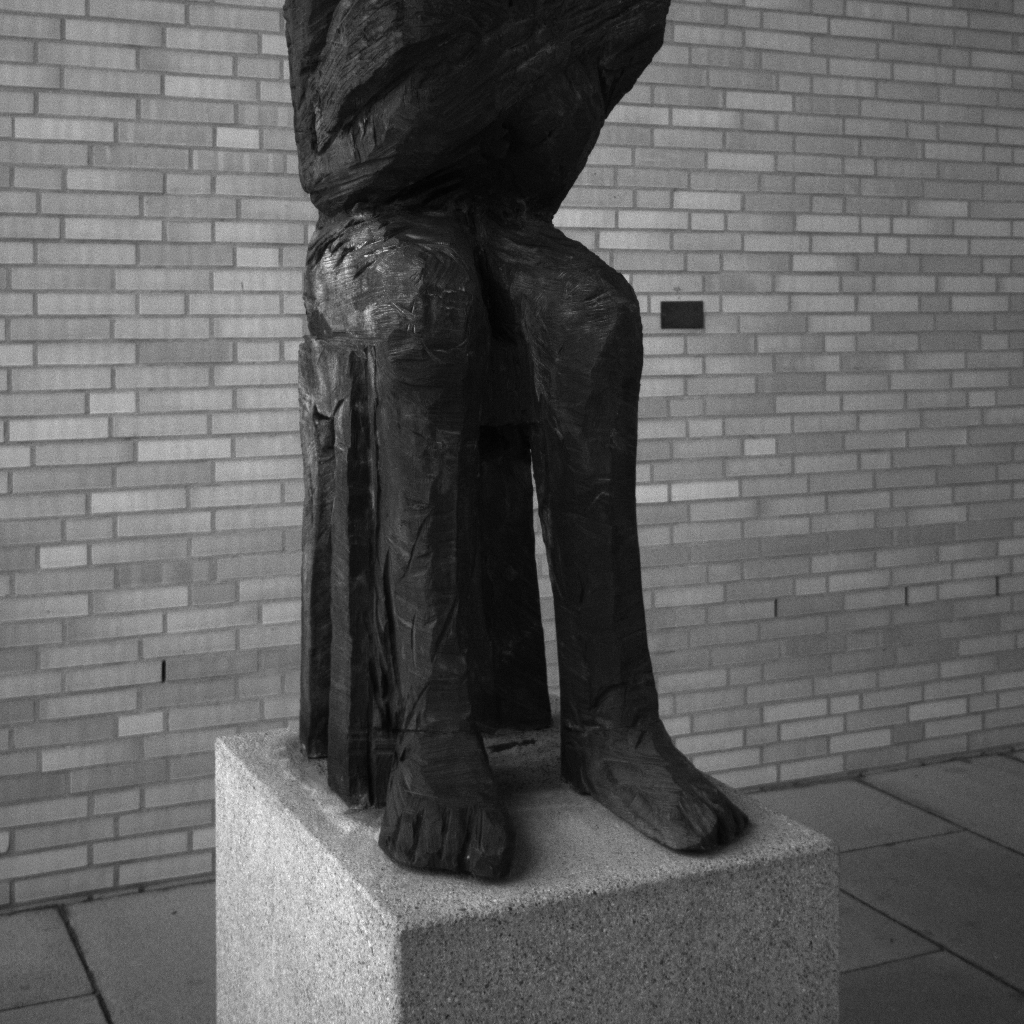}};
  
% img_original_[::2, ::2][700:700+1024, 250:250+1024][450:550, 200:400]

\node at (image.south east) [anchor=south east, inner sep=0] {\includegraphics[width=0.14\textwidth]{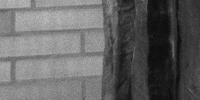}};

\begin{scope}[x={(image.south east)},y={(image.north west)}]
  \begin{scope}[shift={(0,1)},x={(1/1024,0)},y={(0,-1/1024)}]
  
   \draw[dodgerblue, semithick] (200, 450) rectangle (400, 550);

   \draw[dodgerblue, thick] (424, 724) rectangle (1024, 1024);

    \draw[densely dashed, thin, dodgerblue] (200, 550) -> (424, 1024);
    \draw[densely dashed, thin, dodgerblue] (400, 450) -> (1024, 724);
  \end{scope}
\end{scope}
\end{tikzpicture} \\
\scriptsize Noisy 
\end{tabular}
& 
\begin{tabular}{ccc}
\includegraphics[width=0.225\textwidth]{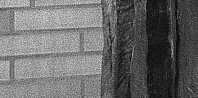} & \includegraphics[width=0.225\textwidth]{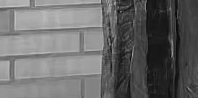} & \includegraphics[width=0.225\textwidth]{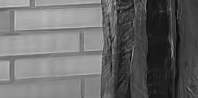} \\
\scriptsize  Noisy & \scriptsize  BM3D \cite{BM3D} & \scriptsize NL-Ridge \cite{nlridge} \\

\includegraphics[width=0.225\textwidth]{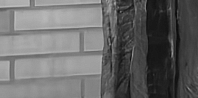} & \includegraphics[width=0.225\textwidth]{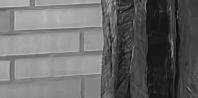} & \includegraphics[width=0.225\textwidth]{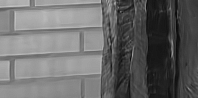} \\
\scriptsize DnCNN \cite{ffdnet}  & \scriptsize  WNNM \cite{WNNM} & \scriptsize LIChI (ours) \\
\end{tabular}

\\
\\

(b)&
\begin{tabular}{c}
\begin{tikzpicture}
\node[anchor=south west,inner sep=0] (image) at (0,0)
  {\includegraphics[width=0.24\textwidth]{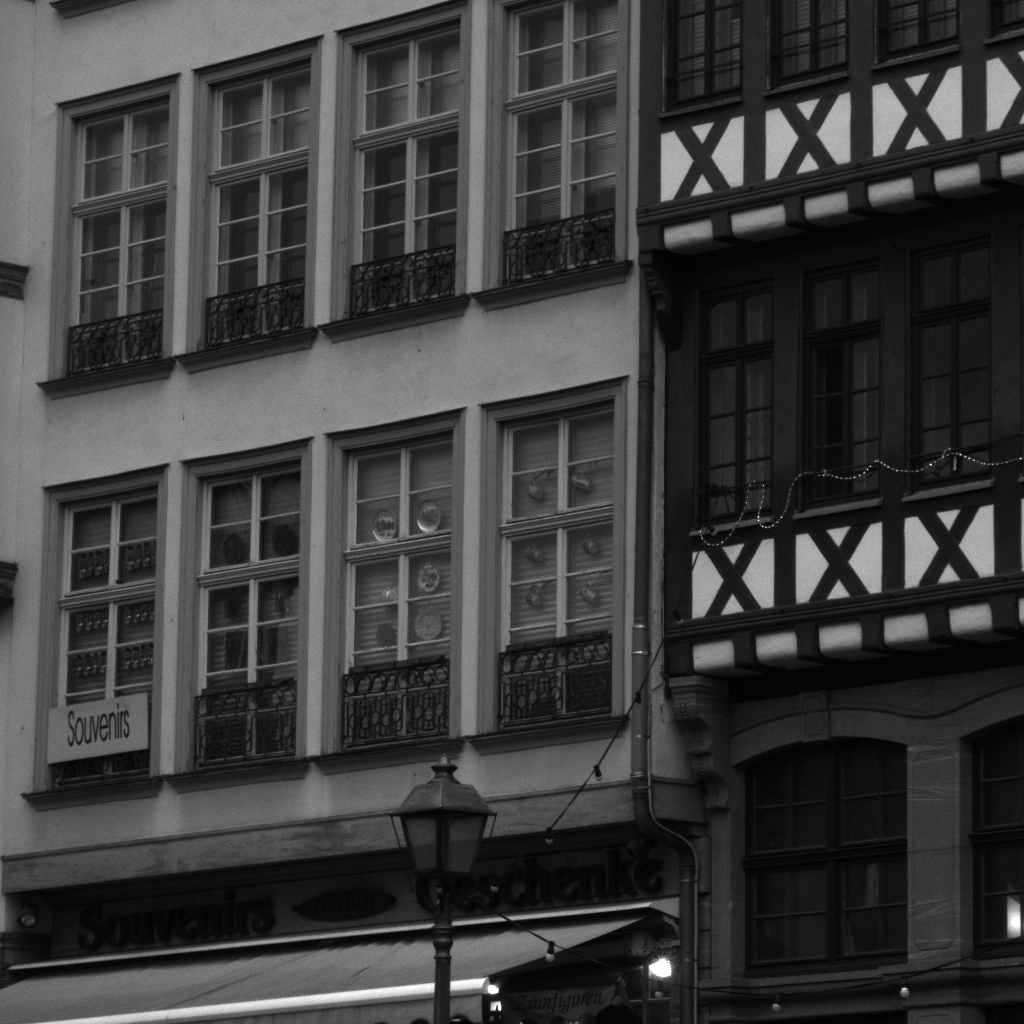}};
  
% img_original_[::2, 1::2][700:700+1024, 250:250+1024][280:380, 200:400]

\node at (image.south east) [anchor=south east, inner sep=0] {\includegraphics[width=0.14\textwidth]{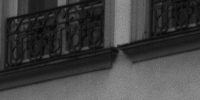}};

\begin{scope}[x={(image.south east)},y={(image.north west)}]
  \begin{scope}[shift={(0,1)},x={(1/1024,0)},y={(0,-1/1024)}]
  
   \draw[dodgerblue, semithick] (200, 280) rectangle (400, 380);

   \draw[dodgerblue, thick] (424, 724) rectangle (1024, 1024);

    \draw[densely dashed, thin, dodgerblue] (200, 380) -> (424, 1024);
    \draw[densely dashed, thin, dodgerblue] (400, 280) -> (1024, 724);
  \end{scope}
\end{scope}
\end{tikzpicture} \\
\scriptsize Noisy 
\end{tabular}
& 
\begin{tabular}{ccc}
\includegraphics[width=0.225\textwidth]{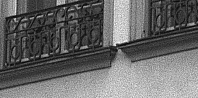} & \includegraphics[width=0.225\textwidth]{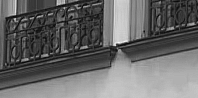} & \includegraphics[width=0.225\textwidth]{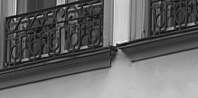} \\
\scriptsize  Noisy & \scriptsize  BM3D \cite{BM3D} & \scriptsize NL-Ridge \cite{nlridge} \\

\includegraphics[width=0.225\textwidth]{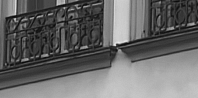} & \includegraphics[width=0.225\textwidth]{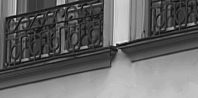} & \includegraphics[width=0.225\textwidth]{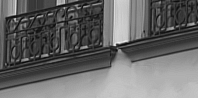} \\
\scriptsize DnCNN \cite{ffdnet}  & \scriptsize  WNNM \cite{WNNM} & \scriptsize LIChI (ours)  \\
\end{tabular}

\\
\\

(c)&
\begin{tabular}{c}
\begin{tikzpicture}
\node[anchor=south west,inner sep=0] (image) at (0,0)
  {\includegraphics[width=0.24\textwidth]{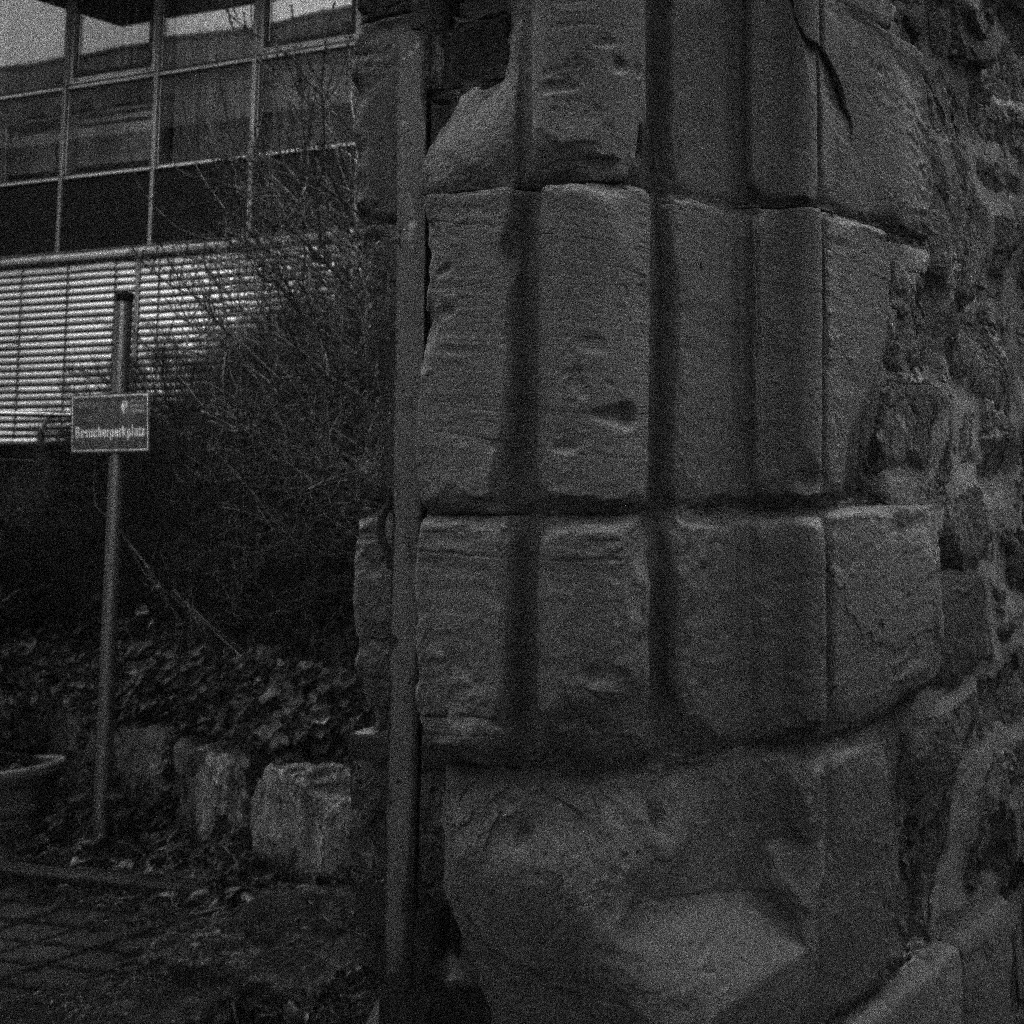}};
  
% img_original_[1::2, ::2][250:250+1024, 1000:1000+1024][50:150, 510:710]

% img_original_[1::2, 1::2][700:1024+700, :1024][360:460, 0:200]
\node at (image.south east) [anchor=south east, inner sep=0] {\includegraphics[width=0.14\textwidth]{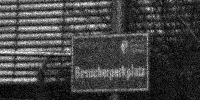}};

\begin{scope}[x={(image.south east)},y={(image.north west)}]
  \begin{scope}[shift={(0,1)},x={(1/1024,0)},y={(0,-1/1024)}]
  
   \draw[dodgerblue, semithick] (0, 360) rectangle (200, 460);
   
   \draw[dodgerblue, thick] (424, 724) rectangle (1024, 1024);

    \draw[densely dashed, thin, dodgerblue] (0, 460) -> (424, 1024);
    \draw[densely dashed, thin, dodgerblue] (200, 360) -> (1024, 724);
  \end{scope}
\end{scope}
\end{tikzpicture} \\
\scriptsize Noisy 
\end{tabular}
& 
\begin{tabular}{ccc}
\includegraphics[width=0.225\textwidth]{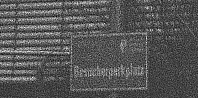} & \includegraphics[width=0.225\textwidth]{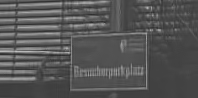} & \includegraphics[width=0.225\textwidth]{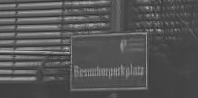} \\
\scriptsize  Noisy & \scriptsize  BM3D \cite{BM3D} & \scriptsize NL-Ridge \cite{nlridge} \\

\includegraphics[width=0.225\textwidth]{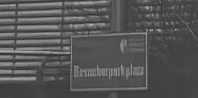} & \includegraphics[width=0.225\textwidth]{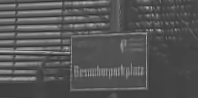} & \includegraphics[width=0.225\textwidth]{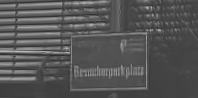} \\
\scriptsize DnCNN \cite{ffdnet}  & \scriptsize  WNNM \cite{WNNM} & \scriptsize LIChI (ours) \\
\end{tabular}

\end{tabular}
\caption{Qualitative comparison of image denoising results on real-world noisy images from Darmstadt Noise Dataset \cite{DND}. Zoom-in regions, enhanced with a $3\times3$ sharpen filter for better visualization,  are indicated for each method. From top to bottom: \textit{Img0003}, \textit{Img0037} and \textit{Img0048}. Best viewed by zooming.}
\label{photo2}
\end{figure*}
\addtolength{\tabcolsep}{5pt}

%% file: conclusion.tex
\section{Conclusion}

In some critical application fields such as biomedical imaging, it is of paramount importance to reduce image noise without creating hallucination artifacts generally induced by the training datasets. It follows that interpretable zero-shot image denoising methods are generally more appreciated by practitioners. Here, we have presented a single-image denoising method based solely on linear combinations of patches that surprisingly achieves state-of-the-art performance. The proposed algorithm is built by extending the unified parametric view of non-local two-step denoisers via a novel chaining rule. Optimization relies on multiple pilot images to guide the estimation of the combination weights. Our experimental results show that LIChI preserves much better structural patterns and textures and generates much less visual artifacts, in particular around the edges, than single-iterated denoisers, such as BM3D. The proposed algorithm is on par with WNNM, the best single-image denoiser to the best of our knowledge,  in terms of both quantitative measurement and visual perception quality, while being much simpler conceptually and faster at execution.

%% file: appendix.tex
\section{Mathematical proofs}
\label{proof_article}

In what follows, $X, Y \in \mathbb{R}^{n \times k}$ with $Y_{i,j} \sim {\cal N}(X_{i,j}, \sigma^2)$ independent along each row and $f_\Theta : Y  \mapsto Y \Theta$  with $\Theta \in \mathbb{R}^{k \times k}$. 

\begin{lemma}
\label{lemma1}
Let $A \in \mathbb{R}^{n \times k}$, $\lambda \in \mathbb{R}^{+} $ and $\mu \in \mathbb{R}$. If  $A^\top A$ is invertible or $\lambda \neq 0$:
\begin{multline*}
\mathop{\arg \min}\limits_{\Theta \in \mathbb{R}^{k\times k}} \; \| A \Theta - A  \|_F^2 + \lambda  \| \Theta \|_F^2 + 2 \mu \operatorname{tr}(\Theta)  \\= I_k  - (\lambda + \mu) (A^\top A + \lambda I_k)^{-1}\,.
\end{multline*}

\begin{proof}
Let $H: \Theta \in \mathbb{R}^{k \times k} \mapsto \| A \Theta - A  \|_F^2 + \lambda  \| \Theta \|_F^2 + 2 \mu \operatorname{tr}(\Theta)$ and $ \displaystyle h_j : \theta \in \mathbb{R}^{k} \mapsto \| A \theta - A_{\cdot, j} \|_2^2 + \lambda \|  \theta \|_2^2 + 2 \mu \theta_{j}$ such that 
$$H(\Theta) = \sum_{j=1}^{k} h_j(\Theta_{\cdot, j})\,.$$

\noindent Note that $\operatorname{Hess} h_j(\theta) = 2(A^\top A + \lambda I_k)$ is symmetric positive definite, hence  $h_j$ is strictly convex and so $h_j$ has a unique global minimum. Then, as $A^\top A + \lambda I_k$ is  $\textit{a fortiori}$ invertible, we have, by canceling the gradient:
\begin{multline*}\nabla h_j(\theta) = 2A^\top(A \theta - A_{\cdot, j}) + 2\lambda \theta + 2 \mu e_j = 0 \\ \Leftrightarrow  \theta = (A^\top A + \lambda I_k)^{-1} (A^\top A_{\cdot, j}  - \mu  e_j)\,,\end{multline*}

\noindent and finally, by concatenating all optimal columns, we get:
\begin{multline*} \arg \min_{\Theta} H(\Theta)  = (A^\top A + \lambda I_k)^{-1} (A^\top A  - \mu  I_k) \\= I_k  - (\lambda + \mu) (A^\top A + \lambda I_k)^{-1}\,.\end{multline*}
\end{proof}
\end{lemma}

\begin{proposition}
Let $(\tau_1, \tau_2) \in \mathbb{R}^2$ with $\tau_1 \neq 0$.
\begin{multline*} \arg \min_{\Theta} \, \mathbb{E} \| f_{\Theta}(X + \tau_1 (Y-X)) -  (X + \tau_2 (Y-X))  \|_F^2   
\; \\= \; I_k - (1 - \frac{\tau_2}{\tau_1}) n(\tau_1\sigma)^2 \left( X^\top X + n(\tau_1\sigma)^2 I_k \right)^{-1}\,.\end{multline*}

\begin{proof}
Let $W = Y-X$ and $\Theta' = (\tau_1 \Theta - \tau_2 I_k)$. 

\noindent By development of the squared Frobenius norm and by linearity of expectation:
\begin{multline*}
\mathbb{E} \| f_{\Theta}(X + \tau_1 W) -  (X + \tau_2 W)  \|_F^2 \\=  \mathbb{E} \left( \| X \Theta - X \|_F^2 + 2 \langle X \Theta - X,  W \Theta'  \rangle_F +  \| W \Theta' \|_F^2  \right)
\\= \| X \Theta - X \|_F^2  +  \mathbb{E}\| W \Theta' \|_F^2\,,
\end{multline*}

\noindent with
\begin{multline*}\mathbb{E}\| W \Theta' \|_F^2 = \mathbb{E} \left( \sum_{i=1}^{n} \sum_{j=1}^{k} \left( \sum_{l=1}^{k} W_{i,l} \Theta'_{l,j} \right)^2 \right)  
\\= \sum_{i=1}^{n} \sum_{j=1}^{k}  \sum_{l=1}^{k} \sigma^2 \Theta^{'2}_{l,j} 
= n\sigma^2  \| \Theta' \|_F^2 \\
= n \sigma^2  \left(\tau_1^2  \| \Theta \|_F^2 - 2 \tau_1 \tau_2 \operatorname{tr}(\Theta)+   k\tau_2^2 \right). 
\end{multline*}

\noindent The use of the Lemma \ref{lemma1} allows to conclude.
\end{proof}
\label{proposition1}
\end{proposition}

\begin{proposition}[Noisier2Noise]
Let $\alpha > 0$ and $y, z$ two vectors with $z \sim \mathcal{N}(y, (\alpha^2 \sigma^2)I)$. We have:
$$\mathbb{E} \left[ \frac{(1+\alpha^2) \phi_{\hat{\boldsymbol{\Theta}}_{\alpha}}(z) - z }{\alpha^2} \right]  = \phi_{\boldsymbol{\Theta}_\alpha^{\textnormal{Nr2N}}}(y)$$
\noindent with $\hat{\boldsymbol{\Theta}}_{\alpha}$ and $\boldsymbol{\Theta}_\alpha^{\textnormal{Nr2N}}$ defined in Section \ref{section_n2n}.

\begin{proof}
First of all, notice that:
$$\frac{(1+\alpha^2) \hat{\Theta}_{\alpha, i} - I_k}{\alpha^2}  =  \Theta_{\alpha, i}^{\textnormal{Nr2N}}\,. $$
Therefore, 
$$\frac{(1+\alpha^2) \phi_{\hat{\boldsymbol{\Theta}}_{\alpha}}(z) - z }{\alpha^2} =  \phi_{\frac{1+\alpha^2}{\alpha^2} \hat{\boldsymbol{\Theta}}_{\alpha} - \frac{1}{\alpha^2} \boldsymbol{I} }(z)= \phi_{\boldsymbol{\Theta}_\alpha^{\textnormal{Nr2N}}}(z)  $$ 
\noindent with $\boldsymbol{I} = \{ I_k \}_{i=1}^{N}$. And finally, by linearity of expectation,
$$\mathbb{E} \left[ \frac{(1+\alpha^2) \phi_{\hat{\boldsymbol{\Theta}}_{\alpha}}(z) - z }{\alpha^2} \right]  = \mathbb{E} \left[ \phi_{\boldsymbol{\Theta}_\alpha^{\textnormal{Nr2N}}}(z)\right] = \phi_{\boldsymbol{\Theta}_\alpha^{\textnormal{Nr2N}}}(y).$$
\end{proof}
\label{proposition3}
\end{proposition}

%% file: acknowlegdment.tex
\section*{Acknowledgment}

This work was supported by Bpifrance agency (funding) through the LiChIE contract. Computations  were performed on the Inria Rennes computing grid facilities partly funded by France-BioImaging infrastructure (French National Research Agency - ANR-10-INBS-04-07, ``Investments for the future'').

We would like to thank R. Fraisse (Airbus) for fruitful  discussions. 